\documentclass{article}

    \PassOptionsToPackage{numbers, compress}{natbib}

    \usepackage[preprint]{neurips_2025}

\usepackage[utf8]{inputenc} 
\usepackage[T1]{fontenc}    
\usepackage{hyperref}       
\usepackage{url}            
\usepackage{booktabs}       
\usepackage{amsfonts}       
\usepackage{nicefrac}       
\usepackage{microtype}      
\usepackage{xcolor}         
\usepackage{wrapfig}

\usepackage{xcolor}
\usepackage{enumitem}
\usepackage{xspace}
\usepackage{booktabs}
\usepackage{colortbl}
\usepackage{multirow}
\usepackage{makecell}
\usepackage{lipsum} 
\usepackage{graphicx}
\usepackage{algorithmic}
\usepackage{algorithm}
\usepackage{titletoc}

\newcommand{\eref}[1]{Eq.~\ref{#1}}

\usepackage[labelsep=period]{caption}
\renewcommand{\paragraph}[1]{\vspace{0.2em}\noindent \textbf{#1 \hspace{0.2em}}}

\definecolor{MyDarkRed}{rgb}{0.66, 0.16, 0.16}
\definecolor{MyDarkBlue}{rgb}{0.16, 0.16, 0.66}

\newcommand{\calb}[1]{\boldsymbol{\mathcal{#1}}}
\newcommand{\bs}[1]{{\boldsymbol{#1}}}

\newcommand{\tex}{\mathcal{T}}
\newcommand{\oritex}{\hat{\mathcal{T}}_0}
\newcommand{\refuv}{\tex_{\mathcal{UV}}}
\newcommand{\blendcoeff}{\lambda}
\newcommand{\latent}{\boldsymbol{z}}

\definecolor{Red}{rgb}{0.6,0,0}
\definecolor{Blue}{rgb}{0,0,0.8}
\definecolor{Green}{rgb}{0.2,0.8,0}
\definecolor{airforceblue}{rgb}{0.36, 0.54, 0.66}
\definecolor{ao(english)}{rgb}{0.0, 0.5, 0.0}
\definecolor{azure(colorwheel)}{rgb}{0.0, 0.5, 1.0}
\definecolor{crimson}{rgb}{0.86, 0.08, 0.24}
\definecolor{darkcerulean}{rgb}{0.03, 0.27, 0.49}
\definecolor{cobalt}{rgb}{0.0, 0.28, 0.67}
\definecolor{rosegold}{rgb}{0.72, 0.43, 0.47}
\definecolor{orange-red}{rgb}{1.0, 0.27, 0.0}
\definecolor{mountainmeadow}{rgb}{0.19, 0.73, 0.56}
\definecolor{malachite}{rgb}{0.04, 0.85, 0.32}
\definecolor{darkblue}{rgb}{0.0, 0.0, 0.55}
\definecolor{customblue}{rgb}{0.2, 0.35, 0.8}

\hypersetup{colorlinks,linkcolor={blue},citecolor={mountainmeadow},urlcolor={magenta}}

\usepackage{amsmath,amsfonts,bm}

\def\eqref#1{equation~\ref{#1}}

\def\1{\bm{1}}

\DeclareMathAlphabet{\mathsfit}{\encodingdefault}{\sfdefault}{m}{sl}
\SetMathAlphabet{\mathsfit}{bold}{\encodingdefault}{\sfdefault}{bx}{n}

\title{Tex4D: Zero-shot 4D Scene Texturing with Video Diffusion Models}

\author{%
\hspace{-0.5cm}
Jingzhi Bao\textsuperscript{\rm 1}\quad 
Xueting Li\textsuperscript{\rm 2}\quad
Ming-Hsuan Yang\textsuperscript{\rm 3}
\\
\hspace{-0.6cm}
\textsuperscript{\rm 1}CUHK-Shenzhen
\quad\textsuperscript{\rm 2}NVIDIA
\quad\textsuperscript{\rm 3}UC Merced
\\
{\tt \normalsize \href{https://tex4d.github.io}{https://tex4d.github.io}}
}

\begin{document}

\maketitle

\vspace{-20pt}

\begin{figure}[!htb]
    \centering
    \includegraphics[width=\linewidth]{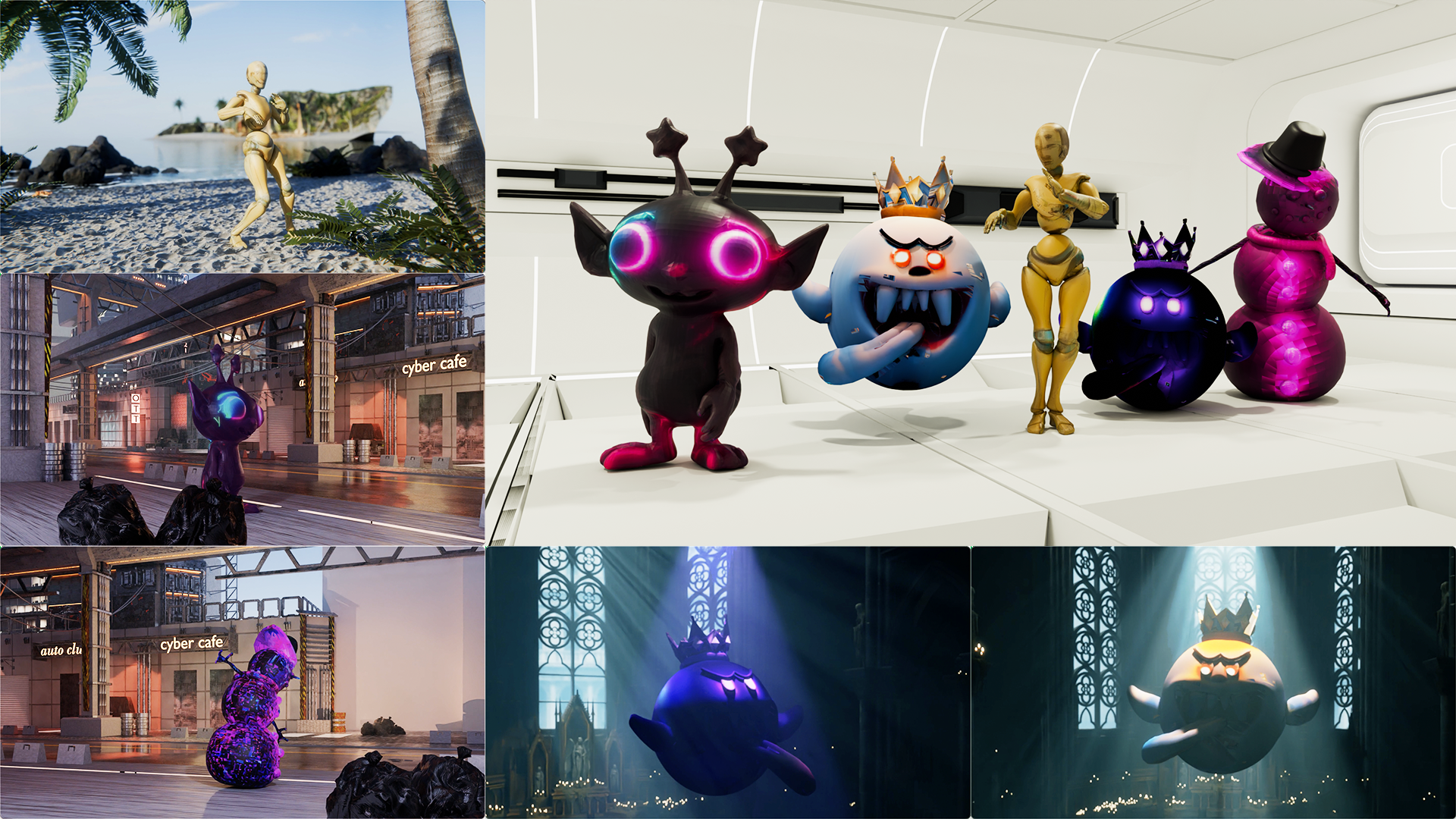}
    \caption{\textbf{Tex4D Application.} Our synthesized dynamic textures can be easily integrated into graphics pipelines.}
    \label{fig:application}
\end{figure}

\newif\ifdrafting
\draftingtrue
\ifdrafting
    \newcommand{\jz}[1]{\textcolor{cyan}{[JZ: #1]}}
    \newcommand{\XT}[1]{{\color{orange}[XT: #1]}}
\else
    \newcommand{\jz}[1]{}
    \newcommand{\XT}[1]{}
\fi


\begin{abstract}
%
3D meshes are extensively employed in movies, games, AR, and VR for their efficiency in animation and minimal memory footprint, leading to the creation of a vast number of mesh sequences. 
However, creating dynamic textures for these mesh sequences to model the appearance transformations remains labor-intensive for professional artists.
%
%
%
In this work, we present \textbf{Tex4D}, a zero-shot approach that creates multi-view and temporally consistent dynamic mesh textures by integrating the inherent 3D geometry knowledge with the expressiveness of video diffusion models.
%
Given an untextured mesh sequence and a text prompt as inputs, our method enhances multi-view consistency by synchronizing the diffusion process across different views through latent aggregation in the UV space. 
To ensure temporal consistency, such as lighting changes, wrinkles, and appearance transformations, we leverage prior knowledge from a conditional video generation model for texture synthesis.
Straightforwardly combining the video diffusion model and the UV texture aggregation leads to blurry results. We analyze the underlying causes and propose a simple yet effective modification to the DDIM sampling process to address this issue.
Additionally, we introduce a reference latent texture to strengthen the correlation between frames during the denoising process.
To the best of our knowledge, Tex4D is the first method specifically designed for 4D scene texturing. 
Extensive experiments demonstrate its superiority in producing multi-view and multi-frame consistent dynamic textures for mesh sequences.
%
\end{abstract}

\begin{figure*}
    \centering
    \includegraphics[width=0.98\linewidth]{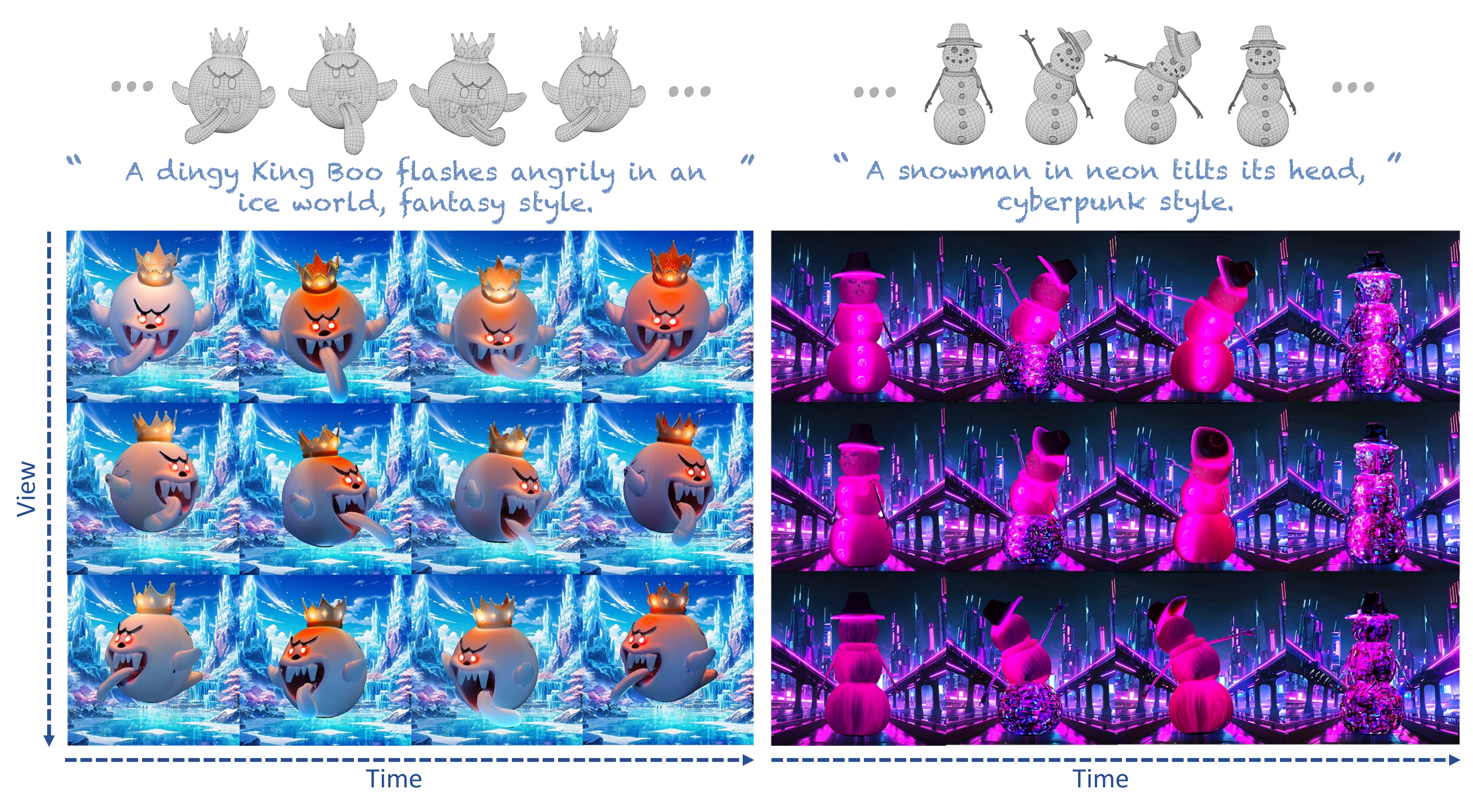}
    \vspace{-.8em}
    \caption{
        Given an untextured mesh sequence and a text prompt as inputs (Top), \textbf{Tex4D} generates multi-view, dynamic textures. Below, we show renderings of the textured meshes from three views and four timestamps.
    }
    \label{fig:teaser}
    \vspace{-1.7em}
\end{figure*}

\section{Introduction}

3D meshes are widely used in modeling, computer-aided design (CAD), animation, and computer graphics due to their low memory footprint and efficiency in animation. 
Visual artists, game designers, and movie creators build numerous animated mesh sequences for visual applications. 
However, creating vivid videos involves complex post-processing steps, such as creating dynamic textures for appearance transformations, as shown in~Fig.~\ref{fig:application}. These steps are labor-intensive and require specialized expertise by artists.

On the other hand, recent advancements in generative models have democratized content creation and demonstrated impressive performance in image and video synthesis. 
For instance, video generation models~\citep{ho2022imagen,esser2023structure,li2023videogen,he2022latent,yu2023video,zhou2022magicvideo,hong2022cogvideo,yang2024cogvideox,2023i2vgenxl,xing2023dynamicrafter,chen2023videocrafter1,chen2024videocrafter2} trained on large-scale video datasets~\citep{Bain21web,schuhmann2021laion} allow users to create realistic video clips from various inputs such as text prompts, images, or geometric conditions.
However, these text-to-video generation models, which are trained solely on 2D data, often struggle with spatial consistency when applied to multi-view image generation~\citep{Tang2023mvdiffusion,shi2023MVDream,liu2023syncdreamer,weng2023consistent123,long2023wonder3d,shi2023zero123++,kwak2023vivid,tang2024mvdiffusionpp,voleti2024sv3d} or 3D object texturing~\citep{cao2023texfusion,liu2023text,richardson2023texture,huo2024texgen}.

To address these limitations, two main approaches have been developed.
One approach~\citep{richardson2023texture,chen2023text2tex,cao2023texfusion} focuses on resolving multi-view inconsistency in static 3D object texturing by synchronizing multi-view image diffusion processes. 
While these methods produce multi-view consistent textures for static 3D objects, they do not address the challenge of generating dynamically changing textures for mesh sequences.
Another approach~\citep{guo2023sparsectrl,lin2024ctrladapter,peng2024controlnext} aims to generate video clips based on the rendering (e.g., depth, normal or UV maps) of an untextured mesh sequence. To encourage temporal consistency, these methods modify the attention mechanism in 2D diffusion models and utilize inherent correspondences in a mesh sequence to facilitate feature synchronization between frames. Although these techniques can be adapted for multi-view image generation by treating camera pose movement as temporal motion, they usually produce inconsistent 3D texturing due to insufficient exploitation of 3D geometry priors. 

In this paper, we introduce a novel task: 4D scene texturing. Given an animated untextured 3D mesh sequence and a text prompt, our goal is to generate dynamic textures that are both temporally and multi-view consistent.
We aim to texture 4D scenes while capturing temporal variations, such as lighting and appearance changes, to produce vivid visual results—a key requirement in downstream tasks like character generation.
Unlike existing works, we fully leverage 3D geometry knowledge from the mesh sequence to enforce multi-view consistency.
Specifically, we develop a method that synchronizes the diffusion process from different views through latent aggregation in the UV space. 
%
%
To ensure temporal consistency, we employ prior knowledge from a conditional video generation model for texture sequence synthesis and introduce a reference latent texture to enhance frame-to-frame correlations during the denoising process.
However, naively integrating the UV texture aggregation into the video diffusion process causes the variance shift problem, leading to blurry results.
To resolve this issue, we propose an effective modification to the DDIM~\citep{song2020ddim} sampling process by rewriting the equation. 
%
%
%
Our method is computationally efficient thanks to its zero-shot nature. The textured mesh sequence can be rendered from any camera view, thus supporting various applications in content creation.
%
Our key contributions are:
\begin{itemize}
[leftmargin=*]
\vspace{-0.2cm}
\setlength\itemsep{-.2em}
    \item We present \textbf{Tex4D}, a zero-shot pipeline for generating high-fidelity dynamic textures that are temporally and multi-view consistent, utilizing video diffusion models and mesh sequence controls.
    \item To leverage priors from existing video diffusion models, we develop an effective modification to the DDIM sampling process to address the variance shift issue caused by multi-view texture aggregation and design a background learning module.
    \item We introduce a reference UV blending mechanism to establish correlations during the denoising steps, addressing self-occlusions, and synchronizing the diffusion process in invisible regions.
    \item Our method is not only computationally efficient, but also demonstrates comparable if not superior performance to various state-of-the-art baselines. 
\end{itemize}

\section{Related Work}


\textbf{Video Stylization and Editing.} Video diffusion models have shown remarkable performance in the field of video generation. These models learn motions and dynamics from large-scale video datasets using 3D-UNet to create high-quality, realistic, and temporally coherent videos. 
Although these approaches show compelling results, the generated videos lack fine-grained control, inhibiting their application in stylization and editing. To solve this issue, inspired by ControlNet~\cite{zhang2023adding}, SparseCtrl~\cite{guo2023sparsectrl} trains a sparse encoder from scratch using frame masks and sparse conditioning images as input to guide the video diffusion model. CTRL-Adapter~\cite{lin2024ctrladapter} proposes a trainable intermediate adapter to connect the features between ControlNet and video diffusion models.

Meanwhile, \cite{Tumanyan2023pnp} observed that the spatial features of T2I models play an influential role in determining the structure and appearance, Text2Video-Zero~\cite{text2video-zero} uses a frame-warping method to animate the foreground object by T2I models and \cite{wu2023tune,ceylan2023pix2video,qi2023fatezero} propose utilizing self-attention injection and cross-frame attention to generate stylized and temporally consistent video using DDIM inversion~\cite{song2020ddim}. 
Subsequently, numerous works~\cite{zhang2023controlvideo,cai2023genren,yang2023rerender,geyer2023tokenflow,eldesokey2024latentman} generate temporally consistent videos utilizing T2I diffusion models by spatial latent alignment without training. However, the synthesized videos usually show flickerings due to the empirical correspondences, such as feature embedding distances and UV maps, which are insufficient to express the continuous content in the latent space. Another line of work~\cite{singer2022make,bar2022text2live,blattmann2023videoldm,xu2023magicanimate,guo2023animatediff} is to train additional modules on large-scale video datasets to construct feature mappings, for example, Text2LIVE~\cite{bar2022text2live} applies test-time training with the CLIP loss, and MagicAnimate~\cite{xu2023magicanimate} introduced an appearance encoder to retain intricate clothes details. 


\textbf{Texture Synthesis.}
With the rapid development of foundation models, researchers have focused on applying their generation capability and adaptability to simplify the process of designing textures and reduce the expertise required.
To incorporate the result 3D content with prior knowledge, earlier works~\cite{khalid2022clipmesh,text2mesh,chen2022tango} jointly optimize the meshes and textures from scratch with the simple semantic loss from the pre-trained CLIP~\cite{radford2021learning} to encourage the 3D alignment between the generated results and the semantic priors. However, the results show apparent artifacts and distortion because the semantic feature cannot provide fine-grained supervision during the generation of 3D content.

DreamFusion~\cite{poole2022dreamfusion} and similar models~\cite{lin2023magic3d,wang2023prolificdreamer,po2024compositional,metzer2022latent,chen2023scenetex} distill the learned 2D diffusion priors from the pre-trained diffusion models~\cite{rombach2021highresolution} to synthesize the 3D content by Score Distillation Sampling (SDS). 
These methods render 2D projections of the 3D asset parameters and compare them against reference images, iteratively refining the 3D asset parameters to minimize the discrepancy of the target distribution of 3D shapes learned by the diffusion model. 
Although these approaches enable people without expertise to generate detailed 3D content by textual prompt, their results are typically over-saturated and over-smoothed, hindering their application in actual cases.
Another line of optimization-based methods~\cite{yu2023texture,zeng2024paint3d,bensadoun2024meta} turned to fuse 3D shape information, such as vertex positions, depth maps, and normal maps, with the pre-trained diffusion model by training separate modules on 3D datasets. Still, they require a specific UV layout process to achieve plausible results.

Recently, TexFusion~\cite{cao2023texfusion} and numerous zero-shot methods~\cite{liu2023text,richardson2023texture,huo2024texgen} have shown significant success in generating globally consistent textures without additional 3D datasets. Based on depth-aware diffusion models, they sequentially inpaint the latents in the UV domain to ensure the spatial consistency of latents observed across different views. Then, they decode the latents from multiple views and finally synthesize the RGB texture through back projection.

However, these methods generate static 3D assets and overlook temporal changes in visual presentations, such as videos. To our knowledge, this is the first approach to synthesize multi-view dynamic textures for mesh sequences, enabling appearance transformations.
\section{Preliminaries}
\label{sec:prelim}


\textbf{Video Diffusion Prior.} In this paper, we adopt CTRL-Adapter \citep{lin2024ctrladapter} as our prior model to provide dynamic information. 
CTRL-Adapter aims to adapt a pre-trained text-to-video diffusion model to conditions for various types of images, such as depth or normal map sequences. 
The key idea behind CTRL-Adapter is to leverage a pre-trained ControlNet~\citep{zhang2023adding} and to align its latents with those of the video diffusion model through a learnable mapping module. 
Intuitively, the video diffusion model generates temporally consistent video frames that capture dynamic elements like character motions and lighting, while the ControlNet further enhances this capability by allowing the model to condition on geometric information, such as depth and normal map sequences. 
This makes CTRL-Adapter particularly effective in providing a temporally consistent texture prior to our 4D scene texturing task.
Specifically, we leverage the depth-conditioned CTRL-Adapter model. Given a sequence of depth images denoted as $\{D_1, ..., D_K\}$ and a text prompt $\calb{P}$, CTRL-Adapter (denoted as $\calb{C}$) synthesizes a frame sequence $F$ by $F = \calb{C}(\{D_1, ..., D_K\}, \calb{P})$.
%
%
%

\textbf{DDIM Sampling.} DDIM~\citep{song2020ddim} is a widely used sampling method in diffusion models due to its superior efficiency and deterministic nature compared to DDPM~\citep{ho2020denoising}. 
To enhance numerical stability and prevent temporal color shifts in video diffusion, numerous models~\citep{2023i2vgenxl,ho2022imagen} employ a learning-based sampling technique known as v-prediction~\citep{salimans2022progressive}. 
%
At each denoising step, the sampling process for the latents (denoted as $\latent_t$) can be described as follows:
\begin{equation}\label{eq:ddim}
\begin{aligned}
    \latent_{t-1}=\sqrt{\alpha_{t-1}} \cdot \hat{\latent}_0 (\latent_t) +\sqrt{1-\alpha_{t-1}} \cdot \bs{\epsilon}_\theta (\latent_t),
    \\
    \hat{\latent}_0 (\latent_t) =\frac{\latent_t-\sqrt{1-\alpha_t} \cdot \bs{\epsilon}_\theta}{\sqrt{\alpha_t}},
    \quad
    \bs{\epsilon}_\theta (\latent_t) =\bs{\epsilon}_\theta,
\end{aligned}
\end{equation}
where $\alpha_t$ is the noise variance at time step $t$, $\bs{\epsilon}_\theta$ is the estimated noise from the U-Net denoising module, which is expected to follow $\mathcal{N} (0, \mathcal{I})$, and $\hat{\latent}_0 (\latent_t)$ denotes the predicted original sample (i.e., the latents at timestep 0). 
After the v-parameterization, the predicted original sample $\hat{\latent}_0 (\latent_t)$ and the predicted epsilon $\bs{\epsilon}_\theta (\latent_t)$ are computed as follows:
\begin{equation}
\label{eq:v-param}
\begin{aligned}
    \hat{\latent}_0 (\latent_t) = \sqrt{\alpha_{t}} \cdot \latent_t - \sqrt{1-\alpha_t} \cdot \bs{\epsilon}_\theta, \quad
    \bs{\epsilon}_\theta (\latent_t) = \sqrt{\alpha_{t}} \cdot \bs{\epsilon}_\theta + \sqrt{1-\alpha_t} \cdot \latent_t. 
\end{aligned}
\end{equation}
We leverage an enhanced DDIM sampling process in video diffusion models, along with a multi-view consistent texture aggregation mechanism to synthesize 4D textures.
\begin{figure*}[t]
    \centering
    \includegraphics[width=0.96\linewidth]{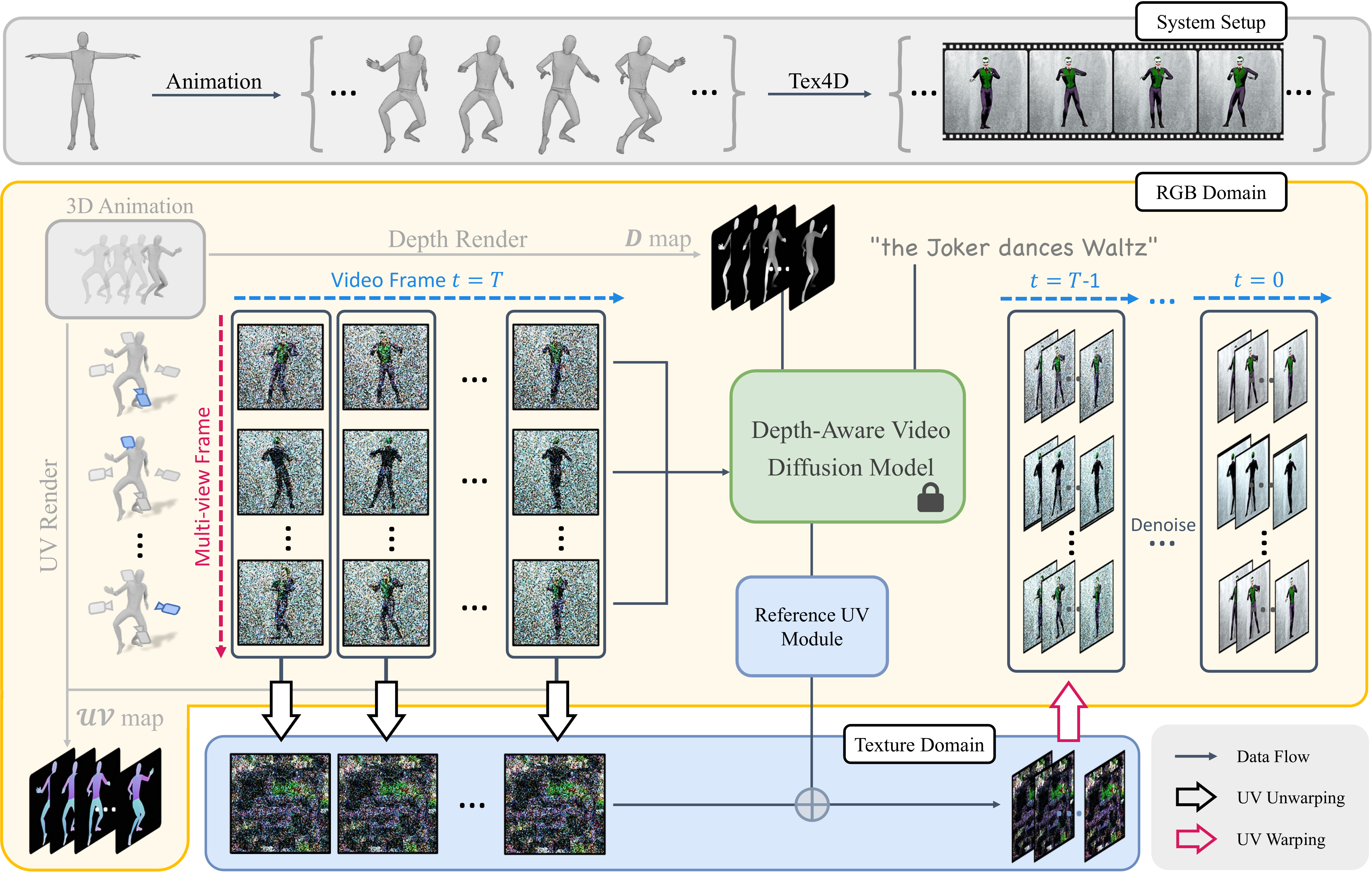}
    \caption{\textbf{Overview.} Given a mesh sequence and a text prompt as inputs, Tex4D generates a UV-parameterized texture sequence that is both globally and temporally consistent.
    At each diffusion step, latent views are aggregated into UV space, followed by multi-view latent texture diffusion to ensure global consistency. To maintain temporal coherence and address self-occlusions, a Reference UV Blending module is applied at each step. Finally, the latent textures are back-projected and decoded to produce RGB textures for each frame.
    }
    \label{fig:method}
    \vspace{-1.5em}
\end{figure*}

\section{Method}


Given an untextured mesh animation and a text prompt, our goal is to generate a multi-view and multi-frame consistent texture sequence for each mesh that aligns with both the text description and motion cues while capturing the dynamics from video diffusion models.

To optimize computational efficiency,
we uniformly sample $K$ key frames from the video and synthesize textures for these keyframes. Textures for the remaining frames are then generated by interpolating the key frame textures.
Formally, given $K$ animated meshes at the keyframes ($\{M_1,...,M_K\})$, along with a text description $\calb{P}$, our method produces 
temporally and spatially consistent UV maps denoted as $\{UV_1, ..., UV_{K}\}$, in a zero-shot manner.

Previous texture generation methods ~\citep{richardson2023texture,chen2023text2tex,cao2023texfusion} typically inpaint and update textures sequentially using pre-defined camera views in an incremental manner. 
However, these approaches rely on view-dependent depth conditions and lack global spatial consistency, often resulting in visible discontinuities in the assembled texture map. This issue arises from error accumulation during the autoregressive view update process, as noted by~\cite{bensadoun2024meta}.
To resolve these issues, rather than processing each view independently, recent methods~\citep{liu2023text} propose to generate multi-view textures simultaneously through diffusion. 
In this work, we similarly leverage the UV space as an intermediate representation to ensure multi-view consistency. 
%
%
\subsection{Overview}
As shown in Fig.~\ref{fig:method}, given a sequence of $K$ meshes, we start by rendering the mesh at $V$ predefined, uniformly sampled camera poses to obtain multi-view depth images (denoted as $\{D_{1,1},...,D_{1,K}, D_{2,1} ..., D_{V, K}\}$), which serve as the geometric conditions.
To generate textures for each mesh, we initialize $V\times K$ noise images sampled from a Normal distribution (denoted as $\{ \latent^{1,1},...,\latent^{1,K}, \latent^{2,1},...,\latent^{V,K} \}$).
Additionally, we initialize an extra noise map sequence $\{\latent_b^{1},...,\latent_b^{K}\}$ for the backgrounds learning. This noise map corresponds to the texture of a plane mesh that is composited with the foreground object at each diffusion step (See Sec.~\ref{sec:muliframe}).
Next, for each view $v\in \{1,...,V\}$, we apply the video diffusion model~\citep{lin2024ctrladapter} discussed in Sec.~\ref{sec:prelim} to simultaneously denoise all latents and obtain multi-frame consistent images as $\{I^{1,v},...,I^{K,v}\}=\calb{C}(\{D_{1,v},...,D_{K, v}\}, \calb{P})$, where $\calb{P}$ is the provided text prompt.
Finally, we un-project and aggregate all denoised multi-view images for each mesh to formulate temporally consistent UV textures.

Applying the video diffusion model independently to each camera view often results in multi-view inconsistencies. Inspired by~\citep{liu2023text,huo2024texgen,zhang2024clay},
we aggregate the multi-view latents of each mesh in the UV space to merge observations across different views at each denoising step, and then render latent from the latent texture to ensure multi-view consistency.
%
Furthermore, we composite the rendered foreground latents with the background latents at each diffusion step (discussed in Sec.~\ref{sec:texaggr}), which is essential to exploit prior in the video diffusion model (see Fig.~\ref{fig:ablation_bg}).
Nonetheless, such a simple aggregation method introduces blurriness in the final results. In Sec.~\ref{sec:muliframe}, we analyze the underlying causes and propose a simple yet effective method to enhance the denoising process.
Additionally, we create a reference UV to handle self-occlusions and further improve temporal consistency in Sec.~\ref{sec:texblend}.
%

\subsection{Multi-view Latents Aggregation in the UV Space}
\label{sec:texaggr}
We describe the aggregation of multi-view latents in the UV space.
%
%
%
%
For frame $k\in \{1,...,K\}$, we aggregate the multi-view latents $\{\latent^{1,k},\dots,\latent^{V,k}\}$ in the UV space by:
\begin{equation}\label{eq:bake}
\tex^k \left( \latent^k \right)= \frac{
        \sum_{v=1}^V \mathcal{R}^{-1}(\bs{z}^{v, k}, c_v) \odot \text{cos}\left(\theta^{v}\right)^\alpha
    }
    {
        \sum_{v=1}^V \text{cos}\left(\theta^{v}\right)^\alpha 
    },
\end{equation}
where $\mathcal{R}^{-1}$ represents the inverse rendering operator that un-projects the latents to the UV space, thus $\mathcal{R}^{-1}(\bs{z}^{v, k}, \bs{c_v})$ produces a partial latent UV texture from view $v$, $\text{cos}(\theta^v)$ is the cosine map buffered by the geometry shader, recording the cosine value between the view direction and the surface normal for each pixel, $\alpha$ is a scaling factor, and $\bs{c_v}$ denotes one of the predefined cameras.
After multi-view latents aggregation, we obtain multi-view consistent latents by rendering the aggregated UV latent map using $\Tilde{\latent}^{v,k} = \mathcal{R}\left(\tex^k ; \bs{c}_{v} \right)$, where $\mathcal{R}$ is the rendering operation. 
%
%
%
%


%

\subsection{Multi-frame Consistent Texture Generation}
\label{sec:muliframe}
The aggregation process discussed above yields multi-view consistent latents $\{\Tilde{\latent}^{v,k}\}$ for the denoising steps.
%
However, this simple aggregation and projection strategy leads to a blurry appearance, as shown in Fig.~\ref{fig:denoise_alg}(b).
This issue arises primarily because the aggregation process depicted in Eq.~\ref{eq:bake} derails the DDIM denoise process. 
%
Specifically, the estimated noise $\bs{\epsilon}_\theta (\latent_t)$ for each step in \eref{eq:ddim} is expected to follow $\mathcal{N} (0, \mathcal{I})$, but \eref{eq:bake} indicates that after aggregating multi-view latents, the expected norm of variance of the noise distribution would be less than $\mathcal{I}$. We denote this as the ``variance shift'' issue caused by the texture aggregation.

%
To resolve this issue, we rewrite the estimated noise $\epsilon_\theta$ as the combination of the t-step latent $\latent_t$ and the estimated latent $\hat{\latent}_0 (\latent_t)$ at step 0. The v-paramaterized predicted epsilon $\bs{\epsilon}_\theta (\latent_t)$ in \eref{eq:v-param} is can be equivalently expressed as:
\begin{equation}
    \begin{aligned}
        \bs{\epsilon_\theta} &=  \left( \sqrt{\alpha_{t}} \cdot \latent_t - \hat{\latent}_0 (\latent_t) \right) / \sqrt{1-\alpha_t}  \\
        \bs{\epsilon}_\theta (\latent_t) 
        &= \sqrt{\alpha_{t}} \cdot \bs{\epsilon_\theta} + \sqrt{1-\alpha_t} \cdot \latent_t \\
        &= \sqrt{\frac{\alpha_{t}}{1-\alpha_t}} \cdot (\sqrt{\alpha_t} \latent_t - \hat{\latent}_0 (\latent_t))  + \sqrt{1-\alpha_t} \cdot \latent_t.
    \end{aligned}
    \label{eq:epsilon}
\end{equation}
%
In practice, we carry out this denoising technique in the UV space. Specifically, we first compute the original texture map (i.e., texture map at step 0, denoted as $\oritex$) by aggregating the predicted original multi-view image latents through Eq.~\ref{eq:bake}. The noisy latent texture map at time step $t$ (denoted as $\tex_t$) can be similarly computed. We denoise in step $t$ by:
\begin{equation}
\label{eq:texdenoise}
\begin{aligned}
    \tex_{t-1} 
    &= \sqrt{\alpha_{t-1}} \cdot \oritex + \sqrt{1-\alpha_{t-1}} \cdot \left( 
        \sqrt{\frac{\alpha_{t}}{1-\alpha_t}} \cdot (\sqrt{\alpha_t} \tex_t - \oritex) 
        + \sqrt{1-\alpha_t} \cdot \tex_t 
    \right).
\end{aligned}
\end{equation}
%
Through experimentation, we observe that background optimization plays a crucial role in fully exploiting the prior within the video diffusion model.
As shown in Fig.~\ref{fig:denoise_alg}(c), using a simple white background produces blurry results.
This may stem from a mismatch between the white-background images and the training dataset, which likely contains fewer such examples, affecting the denoising process.
To resolve this issue, we compute the final latents as the combination of the foreground latent $\Tilde{\latent}_{t-1}$ projected from the aggregated UV latents and the residual background latent $\latent_{b,t-1}$ denoised by diffusion models. 
Specifically, we composite the estimated latents in the $t-1$ step as follows:
\begin{equation}\label{eq:composite_and_render}
    \latent_{t-1} = \Tilde{\latent}_{t-1} \odot \calb{M}_\text{fg} + \latent_{b,t-1} \odot \left( 1 - \calb{M}_\text{fg} \right), \quad
    \Tilde{\latent}_{t-1}, \calb{M}_\text{fg} = \mathcal{R}\left(\tex_{t-1} ; \bs{c}_{v} \right),
\end{equation}
where $\calb{M}_\text{fg}$ represents the foreground mask of the mesh, and $\mathcal{R}$ is the rendering operation. 

To summarize, our diffusion process starts with $K \times (V + 1)$ randomly initialized noise maps sampled (i.e., $\{\latent_T^{1,k},\dots,\latent_T^{V,k}\}$, for foreground, $\{\latent_b^{1},...,\latent_b^{K}\}$ for background) and denoise them into images simultaneously.
At each denoising step $t$ with the key frame $k$, we derive the estimated noises $\{\bs{\epsilon}_{t-1}^{1,k},\dots,\bs{\epsilon}_{t-1}^{V,k}\}$ using the video diffusion model
and calculate
the estimated original latent $\{\hat{\latent}_{0}^{1,k},\dots,\hat{\latent}_{0}^{V,k}\}$ by \eref{eq:ddim}.
%
Then, we use \eref{eq:bake} to aggregate the latents onto UV space.
%
Next, we utilize~\eref{eq:texdenoise} to take the diffusion step in the UV space, and render the synchronized latents $\{\Tilde{\latent}_{t-1}^{1,k},\dots,\Tilde{\latent}_{t-1}^{V,k}\}$ from latent UVs $\{\tex_{t-1}^1,\dots,\tex_{t-1}^K\}$ to ensure multi-view consistency. 
Finally, we composite the denoised latent with the latents at step $t-1$ according to foreground masks by \eref{eq:composite_and_render}.

\subsection{Reference UV Blending}
\label{sec:texblend}

While the video diffusion model ensures temporal consistency for latents from each view, consistency can sometimes diminish after aggregation in the texture domain. 
This issue primarily stems from the view-dependent nature of the depth conditions and the limited resolution of latents, which can lead to distortions when features from different camera angles are combined onto the UV texture. 
Additionally, self-occlusion during animation often results in information loss in invisible regions.

To address these challenges, we propose a reference UV map to enhance correlations between latent textures across frames. 
Specifically, the reference UV map is constructed by sequentially combining latent textures over time, with each new texture filling only the empty texels of the reference UV map.
Each texture is blended using the reference UV $\refuv$ with a mask $\calb{M}_{\mathcal{UV}}$ that labels the visible region:
\begin{equation}\label{eq:blend}
    \tex_t^k = 
    \left(\left(1 - \blendcoeff\right) \cdot \tex_t^k + \blendcoeff \cdot \refuv \right) \odot \calb{M}^k_{\mathcal{UV}}
    + \refuv \odot \left(1 - \calb{M}^k_{\mathcal{UV}}\right)
\end{equation}
where $\blendcoeff$ is the blending weight for the reference UV in the visible region, while the invisible region is simply replaced with the reference texture. We empirically set the blending weight to $0.2$.



\begin{figure*}[t]
    \centering
    \includegraphics[width=\linewidth]{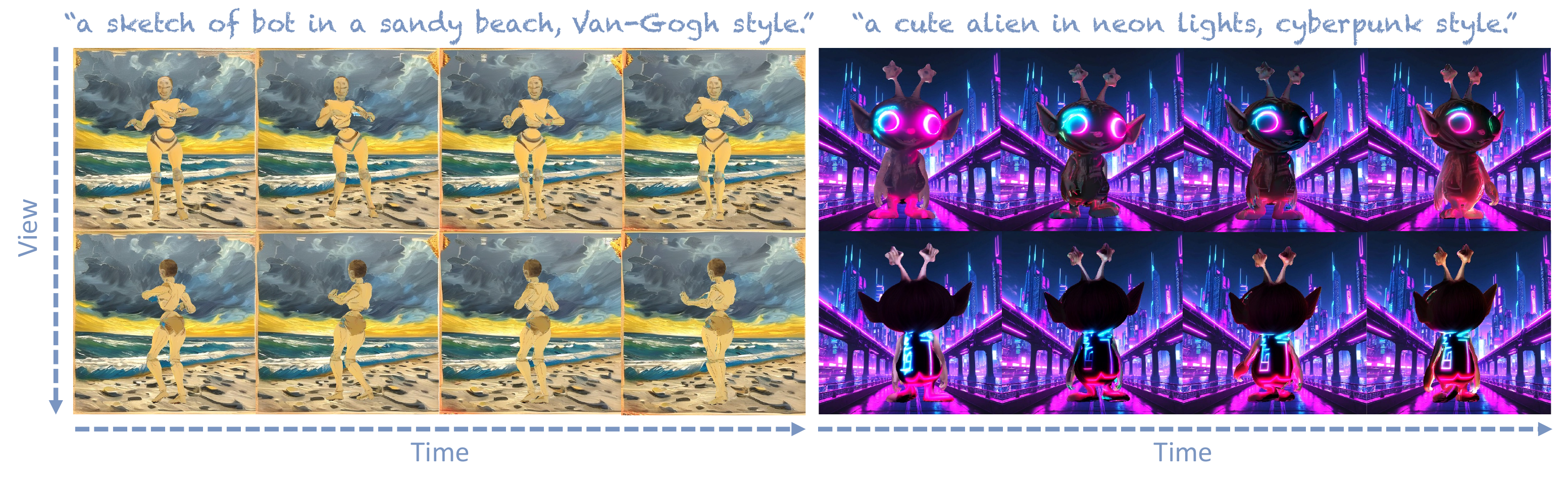}
    \vspace{-2em}
    \caption{\textbf{Qualitative Results.} Our method generates multi-view consistent dynamic textures with a diverse set of styles and prompts. Zoom in to view the details. More results are provided in the supplementary material.}
    \label{fig:qualitative}
    \vspace{-1em}
\end{figure*}

\begin{figure*}[t]
    \centering
    \includegraphics[width=0.96\linewidth]{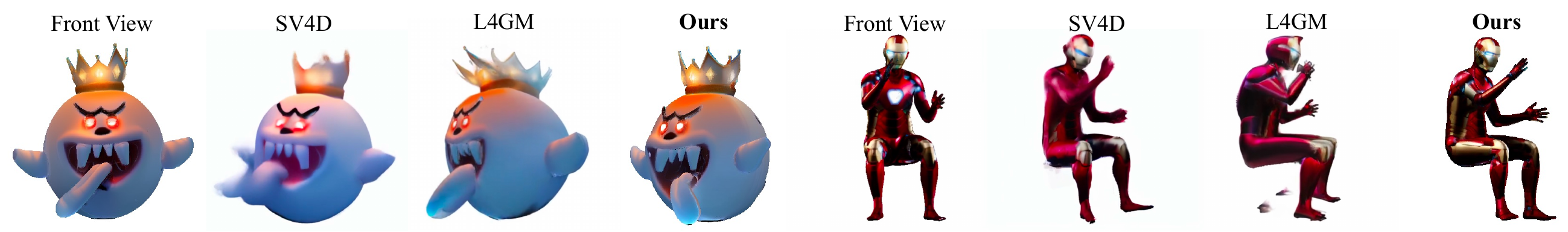}
    \caption{\textbf{Qualitative Comparison with Text-to-4D Methods.} Our methods generates multi-view consistent compared with Text-to-4D methods as our approach fully utilizes the geometry information of the meshes.}
    \label{fig:comp_t24d}
    \vspace{-2em}
\end{figure*}

\section{Experiments}

\textbf{Datasets.} We source our datasets from two primary repositories: human motion diffusion outputs and the Mixamo and Sketchfab websites. We employ the text-to-motion diffusion model~\citep{tevet2023HDM} to compare our approach with LatentMan~\citep{eldesokey2024latentman}.
For comparison with Generative Rendering~\citep{cai2023genren}, we obtain animated characters from the Mixamo. Specifically, we first use Blender~\cite{blender} to extract meshes, joints, skinning weights, and animation data from the FBX files. Then, we apply linear blend skinning to animate the meshes. We utilize XATLAS to parameterize the mesh and unwrap the UVs.

\textbf{Baselines.} To our knowledge, no existing studies tackle dynamic texture generation. We adopt eight recent methods, rendering the input based on their configurations to establish baselines, including Text-to-4D methods, video stylization methods and video generation methods with various control mechanisms.
SV4D~\citep{xie2024sv4d} and L4GM~\citep{ren2024l4gm} are text-to-4D methods, both taking a front view as the input.
PnP-Diffusion~\citep{Tumanyan2023pnp} is an image stylization method that guides the generation with DDIM features.
We extend the method on a frame-by-frame basis for comparison, aligning with previous work~\citep{geyer2023tokenflow}.
Built upon cross-frame attention, Text2Video-Zero~\citep{text2video-zero} guides the video by warping latents to enhance video dynamics implicitly. We leverage its official extension for comparison.
TokenFlow~\citep{geyer2023tokenflow}, Generative Rendering~\citep{cai2023genren}, and LatentMan~\citep{eldesokey2024latentman} establish latent correspondences through nearest neighbor and DensePose features.
Gen-1~\citep{esser2023structure} is a video-to-video model that transforms the untextured mesh renders into stylized outputs.
Given the lack of the source code for Generative Rendering, we utilize the experimental results presented in their video demos for qualitative comparison. 
Additionally, we compare our method with the texture generation method Text2Tex~\citep{chen2023text2tex}.

\textbf{Evaluation Metrics.} Quantitatively evaluating multi-view consistency and temporal coherence remains challenging. We conduct a user study to assess overall performance, including appearance quality, spatio-temporal consistency, and prompt fidelity based on human preferences. Additionally, we measure videl-level multi-view temporal coherence using Fréchet Video Distance (FVD)~\citep{unterthiner2019fvd} following~\citep{li2024vividzoo,xie2024sv4d}, along with a CLIP-based Consistency Score following~\cite{liu2023text}.

\begin{table*}[t]
    \centering
    \captionof{table}{\textbf{Quantitative evaluation}. We present metric values and a comparison of the percentage and statistics of user preference for our approach against other methods. Our method shows the best spatio-temporal consistency as measured by the FVD~\citep{unterthiner2019fvd} and Consistency Score~\citep{liu2023text}. Users consistently favored Tex4D over all baselines.}
    \label{tab:quantitative}

\resizebox{0.9\textwidth}{!}{%
\begin{tabular}{l|ccccc}
\toprule \textbf{Method} & \textbf{FVD ($\downarrow$)} & \textbf{Cons. Score ($\uparrow$)} & \textbf{Appearance Quality} & \textbf{Spatio-temporal Consistency} & \textbf{Consistency with Prompt} \\
\midrule
Text2Video-Zero        & 3078.94 & 86.80 & 89.33\% (2.13) & 91.78\% (2.17) & 91.55\% (2.63) \\
PnP-Diffusion          & 1390.04 & 86.48 & 86.42\% (2.92) & 87.18\% (2.79) & 89.74\% (2.88) \\
TokenFlow              & 1330.43 & 87.35 & 92.31\% (3.08) & 86.84\% (4.04) & 93.42\% (3.08) \\
Gen-1                  & 3114.26 & 81.74 & 70.27\% (4.46) & 75.00\% (3.33) & 77.78\% (4.63) \\
LatentMan              & 2811.23 & 86.50 & 86.57\% (3.63) & 86.57\% (3.88) & 81.82\% (3.75) \\
Ours                   & \textbf{1303.14} & \textbf{95.35} & ~\quad - \quad~ (\textbf{4.69}) & ~\quad - \quad~ (\textbf{4.84}) & ~\quad - \quad~ (\textbf{4.82}) \\
\bottomrule
\end{tabular}%
}



    \vspace{-0.1em}
    \includegraphics[width=0.96\linewidth]{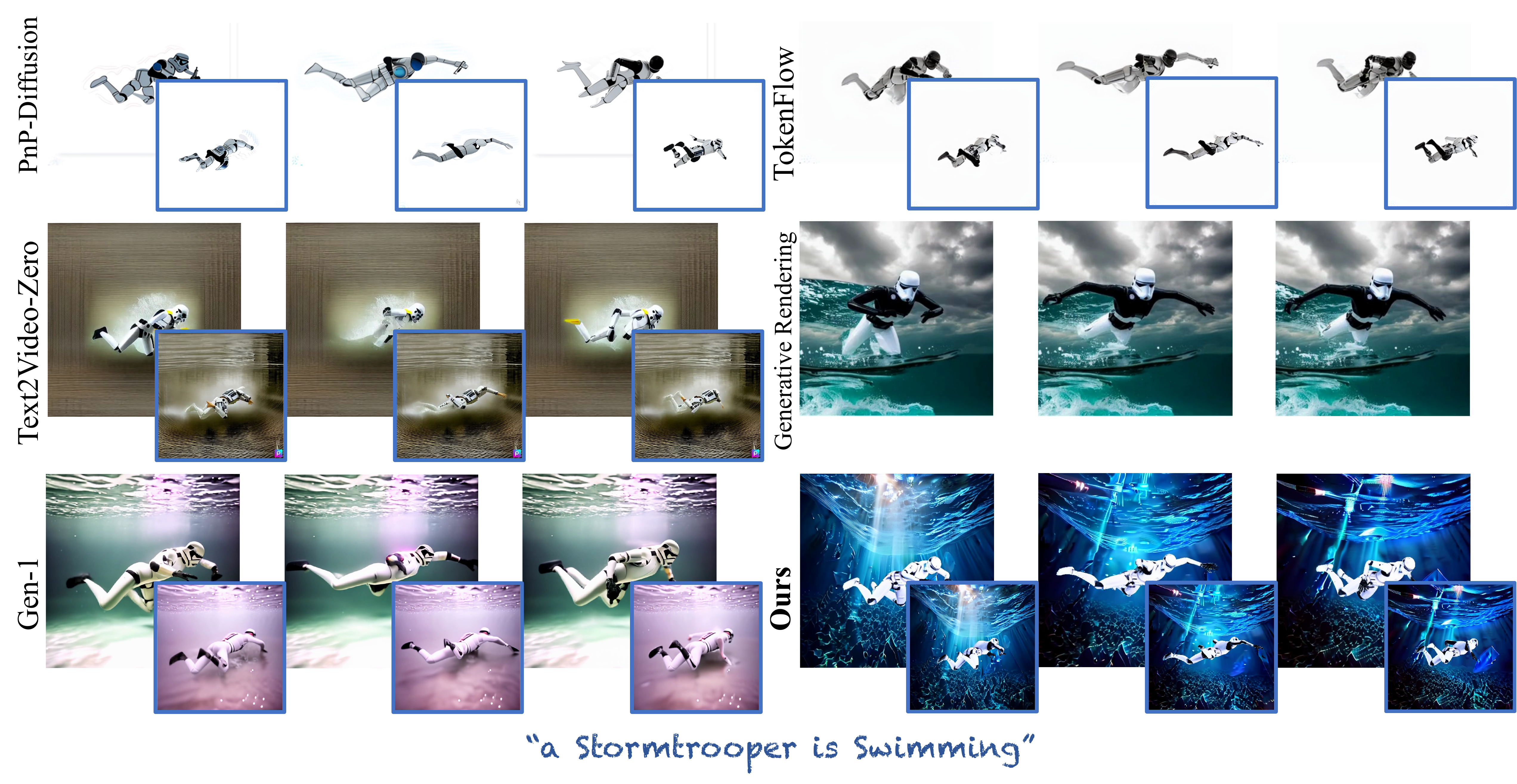}
    \vspace{-0.9em}
    \captionof{figure}{
    \textbf{Qualitative comparisons of multi-view video generation.} We compare our method against PnP-diffusion~\citep{Tumanyan2023pnp}, TokenFlow~\citep{geyer2023tokenflow}, Text2Video-Zero~\citep{text2video-zero}, Generative Rendering~\citep{cai2023genren} (from their video demo), and Gen-1~\citep{esser2023structure}. 
    We generate videos in the front view and the side view (\textcolor{blue}{blue} box) on
    Mixamo dataset. Our method generates vivid videos that align with the textual prompts while preserving spatial consistency.
    }
    \label{fig:comparison}
    \vspace{-2em}
\end{table*}

\subsection{Qualitative Results}

\begin{wrapfigure}{r}{0.47\textwidth}
    \centering
    \vspace{-2em}
    \includegraphics[width=\linewidth]{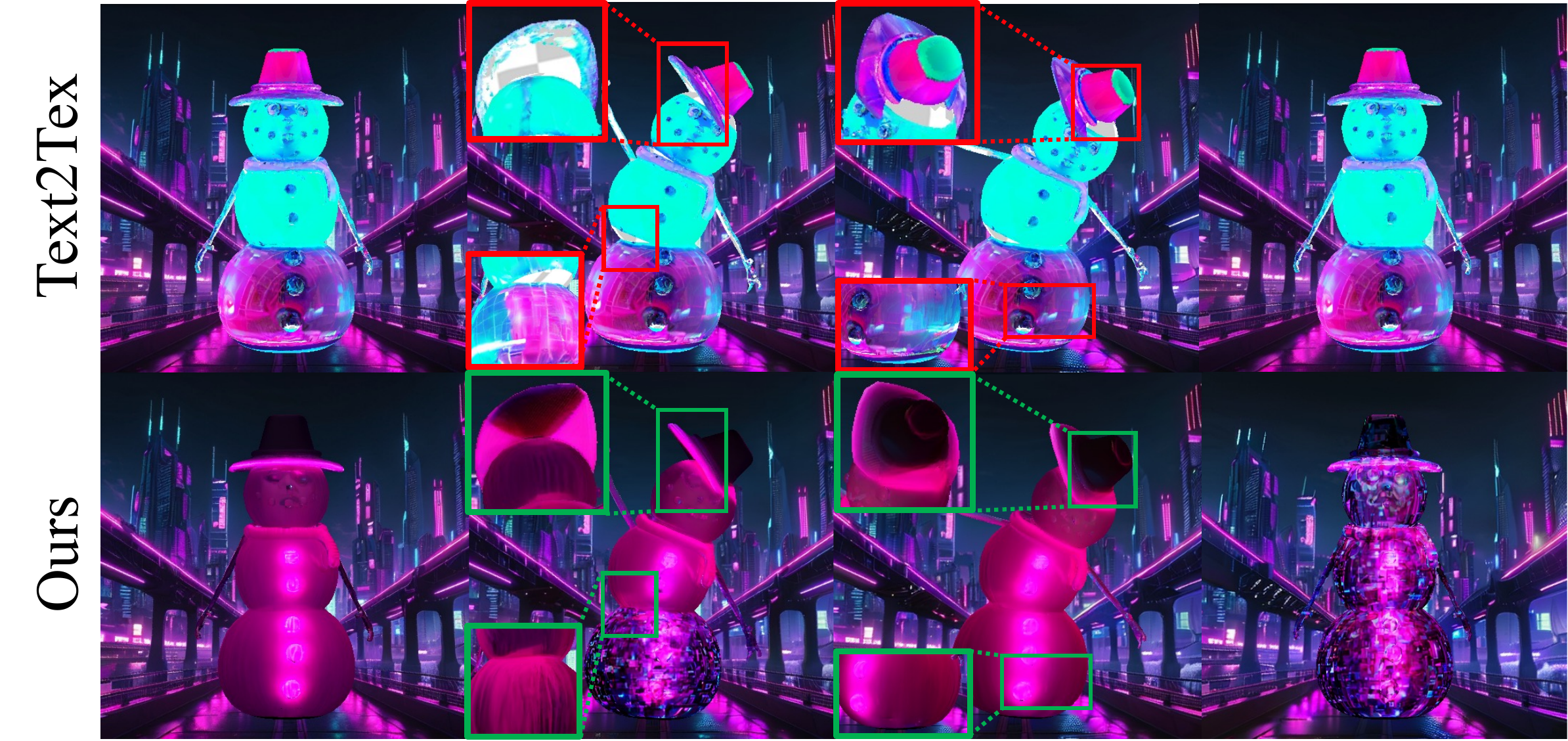}
    \vspace{-1em}
    \caption{\textbf{Comparison with Text2Tex.} Texture generation method~\cite{chen2023text2tex} shows empty texels in invisible regions and fails to model dynamics across time.}
    \label{fig:texcomparison}
    \vspace{-1.5em}
\end{wrapfigure}

We present qualitative evaluation in Fig.~\ref{fig:comp_t24d} and Fig.~\ref{fig:comparison}. 
Text-to-4D methods fail to render plausible results as decoupling geometry and appearance simultaneously is more challenging.
Generative Rendering, TokenFlow, and Text2Video-Zero, which rely on T2I diffusion models with cross-frame attention mechanisms, exhibit noticeable flickering compared to other methods. This issue stems partly from the empirical and implicit correspondence mapping used to enforce interframe latent consistency, as the correspondences in the latent space may not precisely align with those in the RGB space. 
In contrast, our approach interpolates the frames between key frame textures in RGB space, eliminating artifacts caused by latent manipulation.
PnP-Diffusion edits frames independently and generates detailed and sophisticated appearances but suffers from poor spatio-temporal consistency due to the loss of temporal correlations in the latent space.
While Gen-1 produces high-quality videos, it fails to maintain multi-view consistency.

Furthermore, we present multi-view results showcasing a variety of styles and prompts in Fig.~\ref{fig:qualitative}.
Our method, driven by video diffusion models, effectively accounts for the styles and captures temporal variations over time.
As shown in Fig.~\ref{fig:texcomparison}, Tex4D effectively handles the invisible regions compared with the traditional texture generation method Text2Tex~\citep{chen2023text2tex}, which also fails to model dynamics.

\subsection{Quantitative Evaluation}

To quantitatively assess the effectiveness of our proposed method, we follow prior research~\citep{eldesokey2024latentman,geyer2023tokenflow,esser2023structure} and conduct a comprehensive A/B user study. Our study involved 67 participants who provided a total of 1104 valid responses based on six different scenes drawn from six previous works, with each scene producing videos from two different views. During each evaluation, participants were presented with rendered meshes and depth conditions viewed from two angles, serving as motion references. They were shown a pair of videos: one generated by our approach and the other from a baseline method. Participants were asked to select the method that exhibited superior performance in three criteria: 1) appearance quality, 2) spatial and temporal consistency, and 3) fidelity to the prompts. Table~\ref{tab:quantitative} summarizes the preference percentage and statistics (ranging from 1 to 5) of our method over other methods.
Our method significantly surpasses other methods by a large margin.
In addition, our method achieves the best FVD and Consistentcy Score that demonstrates better multi-view consistency in generated video clips.



\begin{figure*}[tp]
    \centering
    \includegraphics[width=\linewidth]{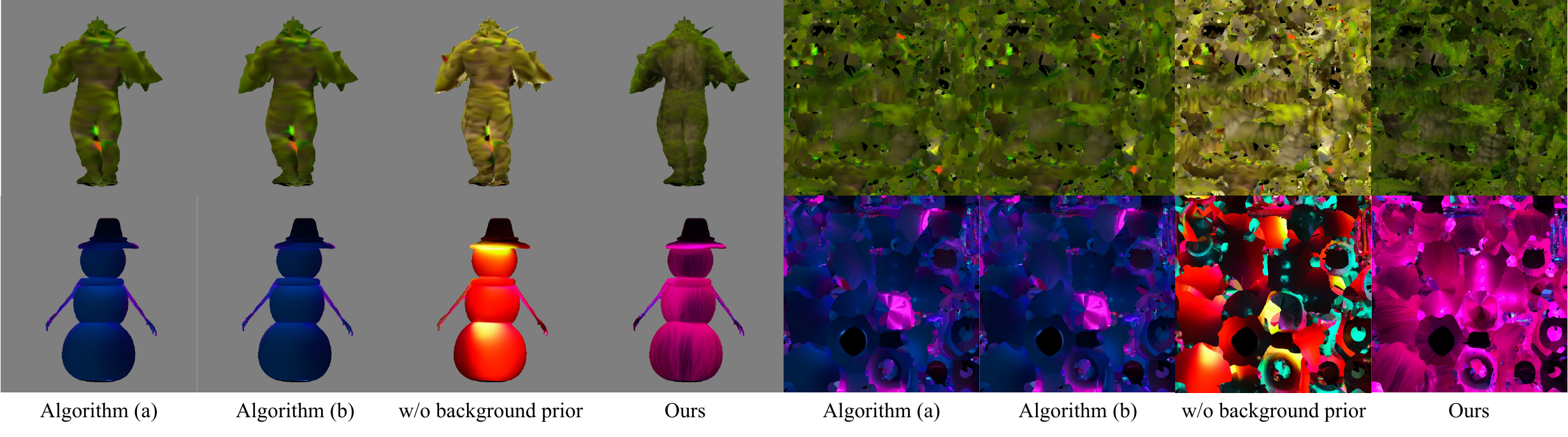}
    \vspace{-1.5em}
    \caption{\textbf{Comparisons of multi-view denoising algorithms and ablation on background priors.} (a) The simple multi-view diffusion algorithms by Eq.~\ref{eq:ddim}, and (b) aggregation of $\latent_{t-1}$ result in a blurry appearance compared to (d) our results. Without background priors, our approach fails to generate plausible video clips.}
    \label{fig:denoise_alg}
    \vspace{-2em}
\end{figure*}

\subsection{Ablation Study}
\label{sec:ablation}

\begin{wrapfigure}{r}{0.44\textwidth}
    \centering
    \vspace{-2.2em}
    \includegraphics[width=\linewidth]{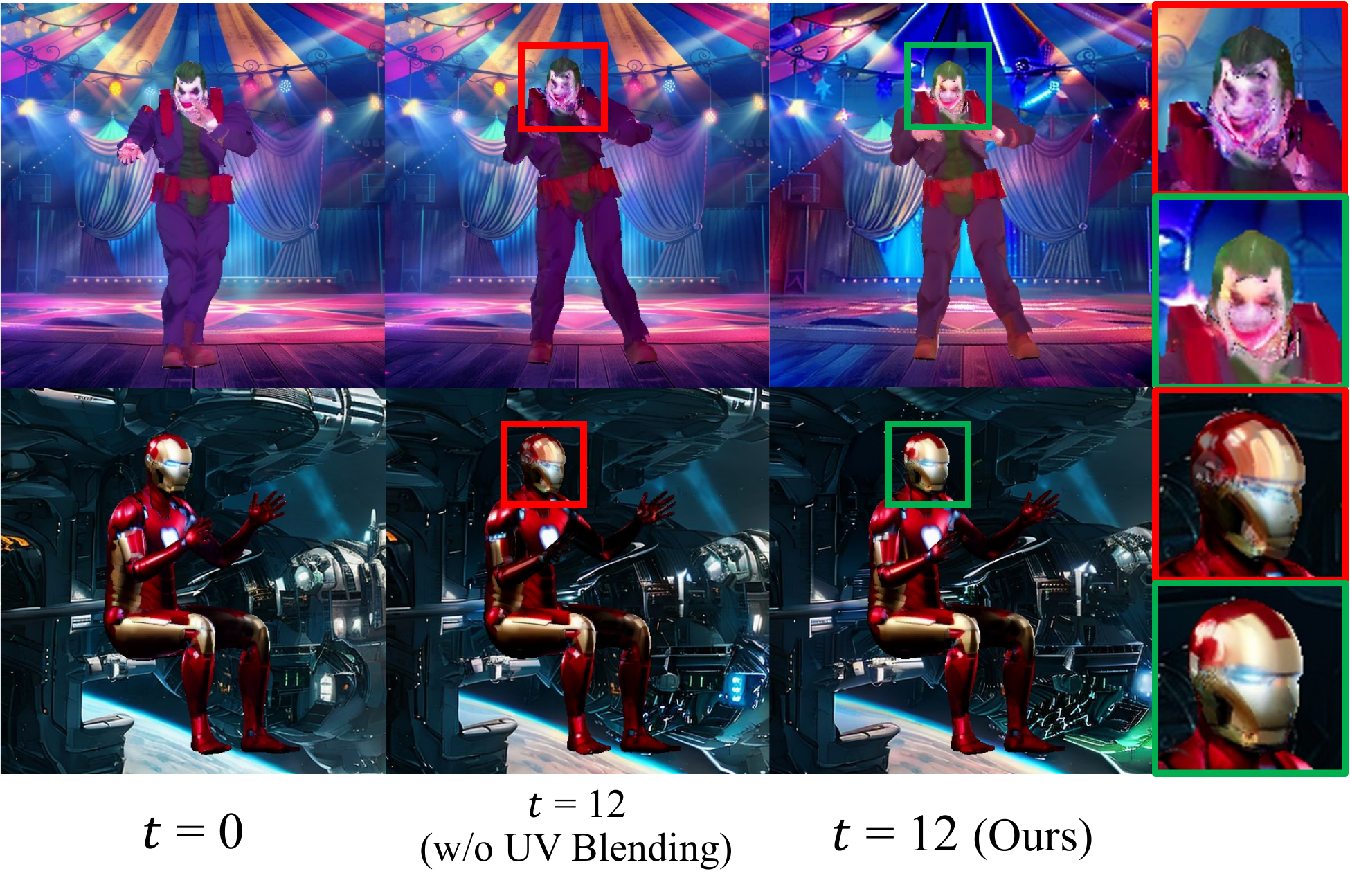}
    \vspace{-1.8em}
    \caption{\textbf{Ablation study on the reference UV blending module.} Without this module, the generated textures lose consistency over time.}
    \label{fig:ablation_blend}
    \vspace{-1.8em}
\end{wrapfigure}

\textbf{Ablation for texture aggregation.} In Fig.~\ref{fig:denoise_alg} (a) and (b), we present two alternative texture aggregation methods. In Fig.~\ref{fig:denoise_alg} (a), we un-project $\hat{\latent}_0 (\latent_t)$ and $\bs{\epsilon}_\theta (\latent_t)$ into UV space for aggregation. In Fig.~\ref{fig:denoise_alg} (b), we map $\latent_{t-1}$ to the UV space. Both these two approaches show inferior results compared to our method, which verifies the effectiveness of the proposed texture aggregation algorithm.

\textbf{Ablation for UV blending module.} 
In Sec.~\ref{sec:texblend}, we propose a reference UV blending schema to resolve the temporal inconsistency caused by latent aggregation.
To validate the effectiveness of this mechanism (See Sec.~\ref{sec:texblend}), we conduct an ablation study by disabling the reference UV blending module (setting $\blendcoeff$ to 0). 
As shown in Fig.~\ref{fig:ablation_blend}, without the UV blending module, our method generates textures with noticeable distortions on the Joker's face over time.

\textbf{Ablation for background priors.} Sec.~\ref{sec:muliframe} discusses the importance of including a plausible background prior.
To verify the effectiveness of this design, we replace the learnable background latents with an all-white background while keeping all other parts unchanged. 
Fig.~\ref{fig:denoise_alg} (c) illustrates that this ablation experiment produces significantly blurrier textures compared to our full method, highlighting the importance of background learning.




\section{Conclusions}
In this paper, we present Tex4D, a zero-shot approach that generates multi-view, multi-frame consistent dynamic textures for untextured, animated mesh sequences based on a text prompt.
By incorporating texture aggregation in the UV space within the diffusion process of a conditional video diffusion model, we ensure both temporal and spatial coherence in the generated textures.
To leverage priors from existing video diffusion models, we develop an effective modification to the DDIM sampling process to address the variance shift issue caused by multi-view texture aggregation and design a background learning module.
Additionally, we enhance temporal consistency by introducing a reference UV map and developing a dynamic background learning framework to produce fully textured 4D scenes.
Extensive experiments show that our method can synthesize realistic and consistent 4D textures, demonstrating its superiority against state-of-the-art baselines.

{
    \small
    \bibliographystyle{abbrv}
    \bibliography{main}
}


\newpage

{
\begin{center}
    \textbf{{{\Large Appendix}}}
\end{center}
}

\appendix
\startcontents[sections]
\printcontents[sections]{l}{1}{\setcounter{tocdepth}{3}}

\newpage

\section{More Implementation Details}


\subsection{Implementation Details}

We utilize the CTRL-Adapter~\citep{lin2024ctrladapter}, trained on the video diffusion model I2VGen-XL~\citep{2023i2vgenxl}, as the backbone for generation, with the denoising steps set to $T=50$. Initially, we center the untextured mesh sequence and pre-define six different viewpoints around the Y-axis in the XZ-plane, uniformly sampled in spherical coordinates, along with an additional top view with an elevation angle of zero and an azimuth angle of $30^\circ$.
For latent initialization, we first sample Gaussian noise on the latent textures and then render 2D latent samples for each view to improve the coherence of the generated outputs. During denoising, we upscale the latent resolution to $96 \times 96$ to reduce aliasing. We empirically set the blending coefficient to $0.2$ during our experiment. It takes approximately 30 minutes to generate a video with 24 keyframes taken on an RTX A6000 Ada GPU. We decode the denoised latents in keyframes to RGB images, and then un-project and aggregate these images to transform the latent UV maps to RGB textures as previous works~\citep{liu2023text,cao2023texfusion,huo2024texgen}. The resolutions of latent texture and RGB texture are $1024\times1024$, $1536\times1536$ respectively. Finally, we interpolate the textures of the keyframes at intervals of 3 to synthesize the final video clips.

\textbf{Background Optimization} For each frame, we use a latent texture as the background shared across multiple views. The first frame is initialized from the image provided by the user as the CTRL-Adapter input. Then, we jointly optimize the background and foreground as described in~Sec. \ref{sec:muliframe}. In our early attempts, we also used a cube or a hemicube mesh to model a free-view background, but found that the backgrounds were not accurate within uncaptured regions by cameras, such as the bottom. Note that our main focus is not a solution for free-view background generation, which is an open research problem explored by concurrent works like Cat4D~\citep{wu2024cat4d} and GenXD~\citep{zhao2024genxd}.

\textbf{Texture Interpolation Details} We compute RGB textures for non-key frames by interpolating RGB textures of nearby keyframes. Specifically, when the denoising stage finishes, we have RGB textures for keyframes, i.e., $T_{r g b, t}=\frac{t-k \cdot n}{n} \cdot T_{r g b, k}+\frac{(k+1) \cdot n-t}{n} \cdot T_{r g b, k+1}$, where $t$ is the $t$-th frame in the final video, $k$ is the nearest prior keyframe, and $n$ is the keyframe interval.

\subsection{Denoising Algorithm of Our Method}

We present the complete workflow of our method in Algorithm~\ref{alg:tex4d_alg}.
For clarity, we omit the notation for the latent variables $\latent_b$ representing the background plane texture, as they follow the same scheme as the foreground latents.
The reference UV map $\refuv$ is constructed by progressively combining latent textures over time, with each new texture filling only the unoccupied texels in the reference UV map. We denote this process as ``$\text{Combine}$'' in the following workflow.

\begin{algorithm}[ht]
    \caption{Tex4D}
    \label{alg:tex4d_alg}
    \vspace{4pt}
    \textbf{Input:}
    {\small
    UV maps $\mathcal{UV}=\{UV_1,...,UV_K\}$;
    depth maps $\calb{D}=\{D_{1,1},...,D_{1,V}, D_{2,1},..., D_{K, V}\}$; 
    text prompt $\calb{P}$;
    CTRL-Adapter model $\calb{C}$;
    rendering operation $\mathcal{R}$;
    unproject operation $\mathcal{R}^{-1}$;
    cameras $\bs{c}$;
    $T$ diffusion steps;
    $\tex$ latent textures (including foreground and background);
    $\blendcoeff$ blending weight;
    $k$ keyframes
    }
    \vspace{6pt}
    
    $\tex_T \sim \mathcal{N}(\bm{0},\mathcal{I})$
    \hfill {\small // Sample noise in UV space}\\
    $\Tilde{\latent}_T, \calb{M}_\text{fg} = \mathcal{R}(\tex_T ; \bs{c})$\\
    $\latent_{b,T} \sim \mathcal{N}(\bm{0},\mathcal{I})$ \\
    $\latent = \latent_T  = \Tilde{\latent}_{T} \odot \calb{M}_\text{fg} + \latent_{b,T} \odot \left( 1 - \calb{M}_\text{fg} \right)$ \hfill {\small // Composite latents} \\
    \textbf{For} $t=T,\dots,1$ \textbf{do}
    \begin{algorithmic}
        \STATE $\latent_{b,t-1} \leftarrow \calb{C}(\latent_{b,t}; \calb{D}, \calb{P}) $  \\
        \STATE $\bs{\epsilon}_\theta \leftarrow \calb{C}(\latent_t; \calb{D}, \calb{P})$ \hfill {\small // Estimate noise from} $\calb{C}$  \\
        \STATE $\hat{\latent}_0 (\latent_t)=\sqrt{\alpha_{t}} \cdot \latent_t-\sqrt{1-\alpha_t} \cdot \bs{\epsilon}_\theta$ \\ 
        \STATE $\oritex, \calb{M}_{\mathcal{UV}} \leftarrow \mathcal{R}^{-1} (\hat{\latent}_0 ; \bs{c}, \mathcal{UV}) $ \hfill {\small // Bake textures by Eq.~\ref{eq:bake}} \\

        \STATE $\refuv = \text{Combine} (\oritex ; \calb{M}_{\mathcal{UV}})$ \\
        \STATE \textbf{For} $k$ in $1, ..., K$ \textbf{do} \\
        \STATE \;\; $\tex_{t-1}^k 
                = \sqrt{\alpha_{t-1}} \cdot \oritex^k 
                + \sqrt{1-\alpha_{t-1}} \left( \sqrt{\frac{\alpha_{t}}{1-\alpha_t}} \cdot (\sqrt{\alpha_t} \tex_t^k - \oritex^k) + \sqrt{1-\alpha_t} \cdot \tex_t^k \right)$ \hfill {\small // Denoise Eq.~\ref{eq:texdenoise}} \\
        
        \STATE  \;\; $\tex_{t-1}^k = 
        \left(\left(1 - \blendcoeff\right) \cdot \tex_{t-1}^k + \blendcoeff \cdot \refuv \right) \odot \calb{M}^k_{\mathcal{UV}}
        + \refuv \odot \left(1 - \calb{M}^k_{\mathcal{UV}}\right)$ \hfill {\small // Blend textures by Eq.~\ref{eq:blend}} \\
        \STATE $ \Tilde{\latent}_{t-1}, \calb{M}_\text{fg} = \mathcal{R}\left(\tex_{t-1} ; \bs{c}, \mathcal{UV} \right)$
        \STATE $\latent_{t-1} = \Tilde{\latent}_{t-1} \odot \calb{M}_\text{fg} + \latent_{b,t-1} \odot \left( 1 - \calb{M}_\text{fg} \right)$ \hfill {\small // Composite latents by Eq.~\ref{eq:composite_and_render}}
        \STATE $\latent = \latent_{t-1}$
        
    \end{algorithmic}
    \textbf{Output:} $\latent$
\end{algorithm}

\subsection{V-Prediction}

Tex4D is a zero-shot approach built on a pre-trained conditional video diffusion model, where v-prediction is a technique commonly used in video diffusion models (e.g., I2VGen-XL~\citep{2023i2vgenxl}, Imagen~\citep{ho2022imagen}, CogVideoX~\citep{hong2022cogvideo,yang2024cogvideox}, CogView3~\citep{zheng2024cogview3}) to accelerate the training and prevent temporal color shifts. In our method, we utilize CTRL-Adapter~\citep{lin2024ctrladapter}, a conditional video diffusion model that guides video by depth maps trained on the DDIM v-prediction mechanism. Hence, we use v-prediction to ensure the proper functioning of the conditional video diffusion model. 


\section{More Qualitative Results}

In this section, we provide more qualitative results of Tex4D to deliver a comprehensive application and analysis of our method. First, we demonstrate the novel application of our method in collaboration with technical artists as in Sec.~\ref{sec:supp_demo}. Our synthesized dynamic texture sequence can be seamlessly integrated into Blender's shader editor and we provide the rendered video demo in the supplementary material. Then, we analyze the multi-view rendering results of Tex4D with a diverse set of prompts and styles in Sec.~\ref{sec:supp_multiview}. Finally, we visualize the generated texture sequence from three different cases to demonstrate that our method can directly bake lighting variations and appearances into textures without the need for post-processing by technical artists in creating vivid videos.

\begin{figure*}
    \centering
    \includegraphics[width=\linewidth]{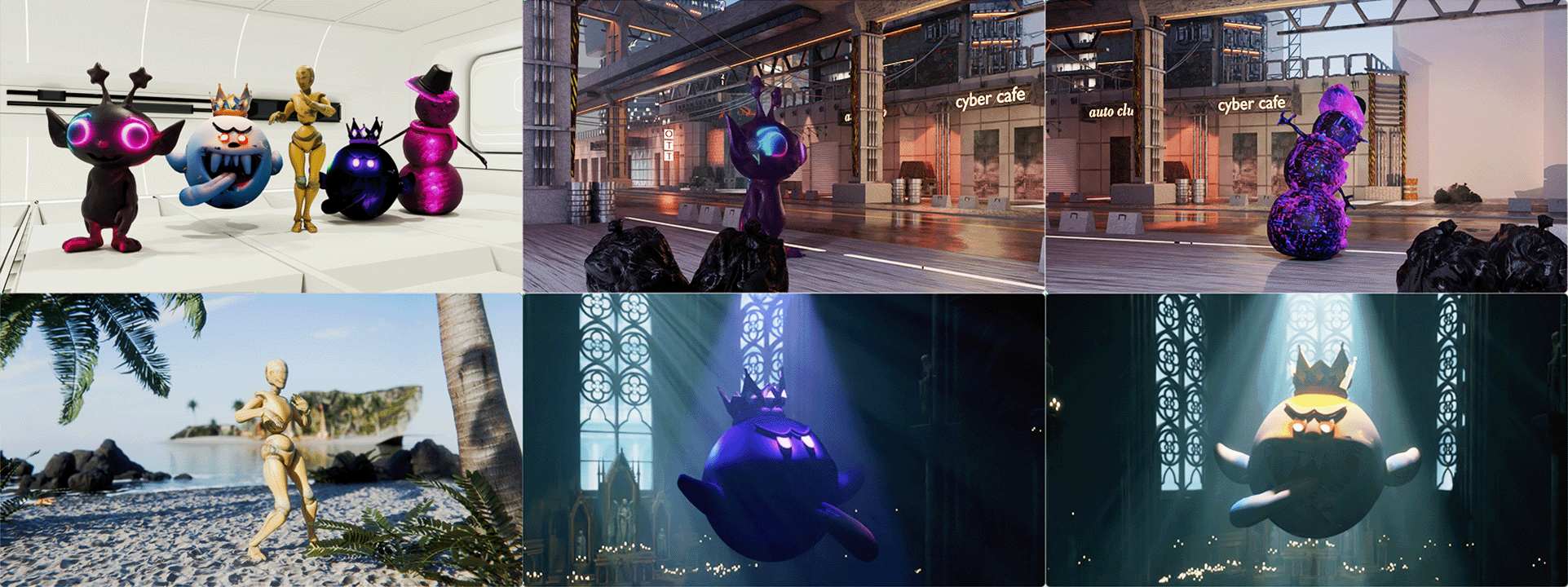}
    \caption{\textbf{Tex4D Applications.} Our synthesized dynamic textures can be easily integrated into graphics pipelines. We utilize the shader editor in Blender to animate textures with image sequence nodes. The dynamic textures help technical artists render vivid videos without additional lighting and mesh controls. We provide a video demo to further demonstrate the effectiveness of Tex4D.}
    \label{fig:tex4d_app}
    \vspace{-1em}
\end{figure*}

\newpage

\begin{figure}[htp]
    \centering
    \includegraphics[width=\linewidth]{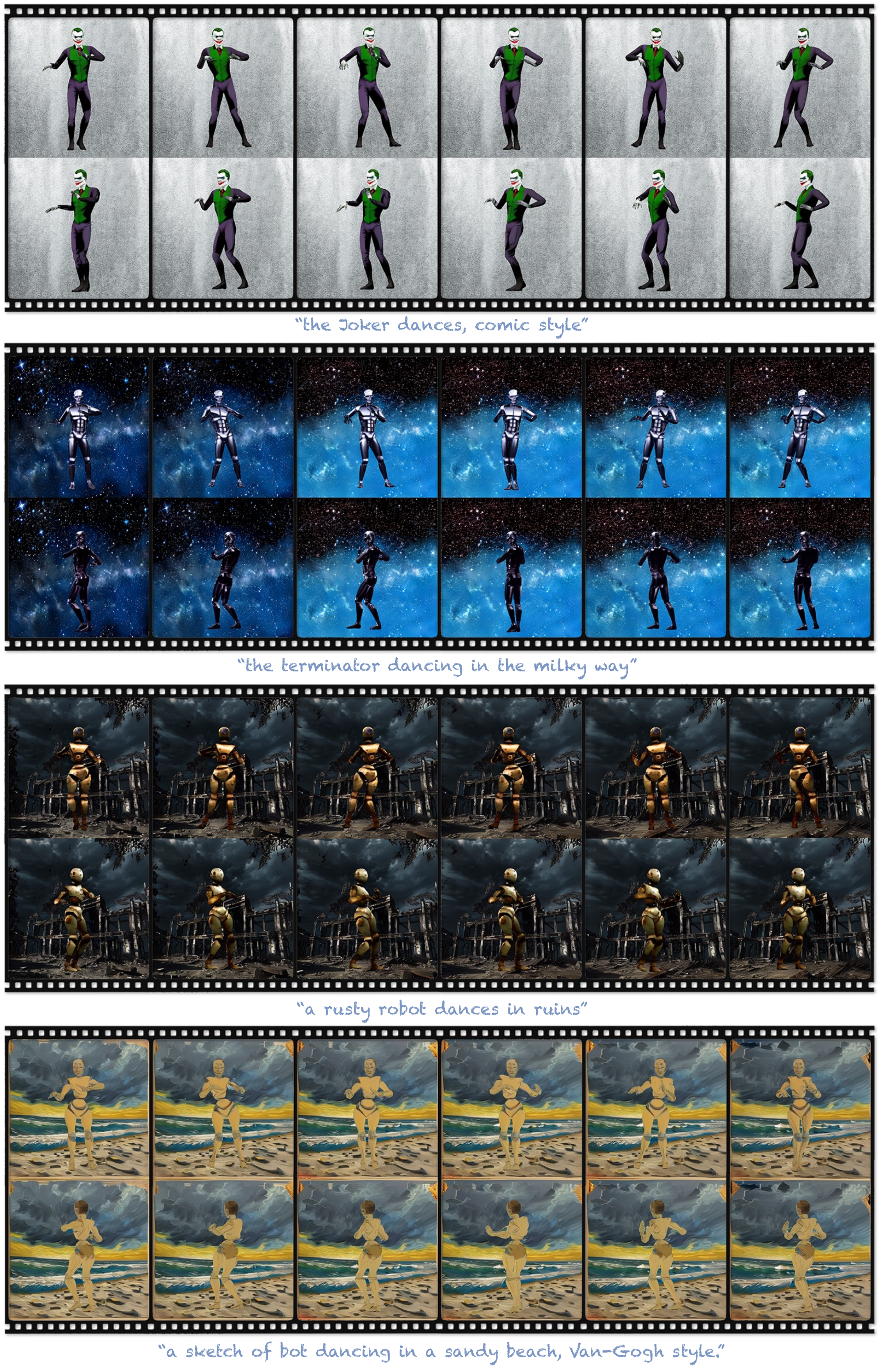}
    \vspace{-1em}
    \caption{\textbf{More qualitative results.} We present the results of Tex4D with brief prompts, demonstrating the ability of Tex4D to generate multi-view consistent textures.}
    \label{fig:supp_results}
\end{figure}

\begin{figure}[htp]
    \centering
    \includegraphics[width=\linewidth]{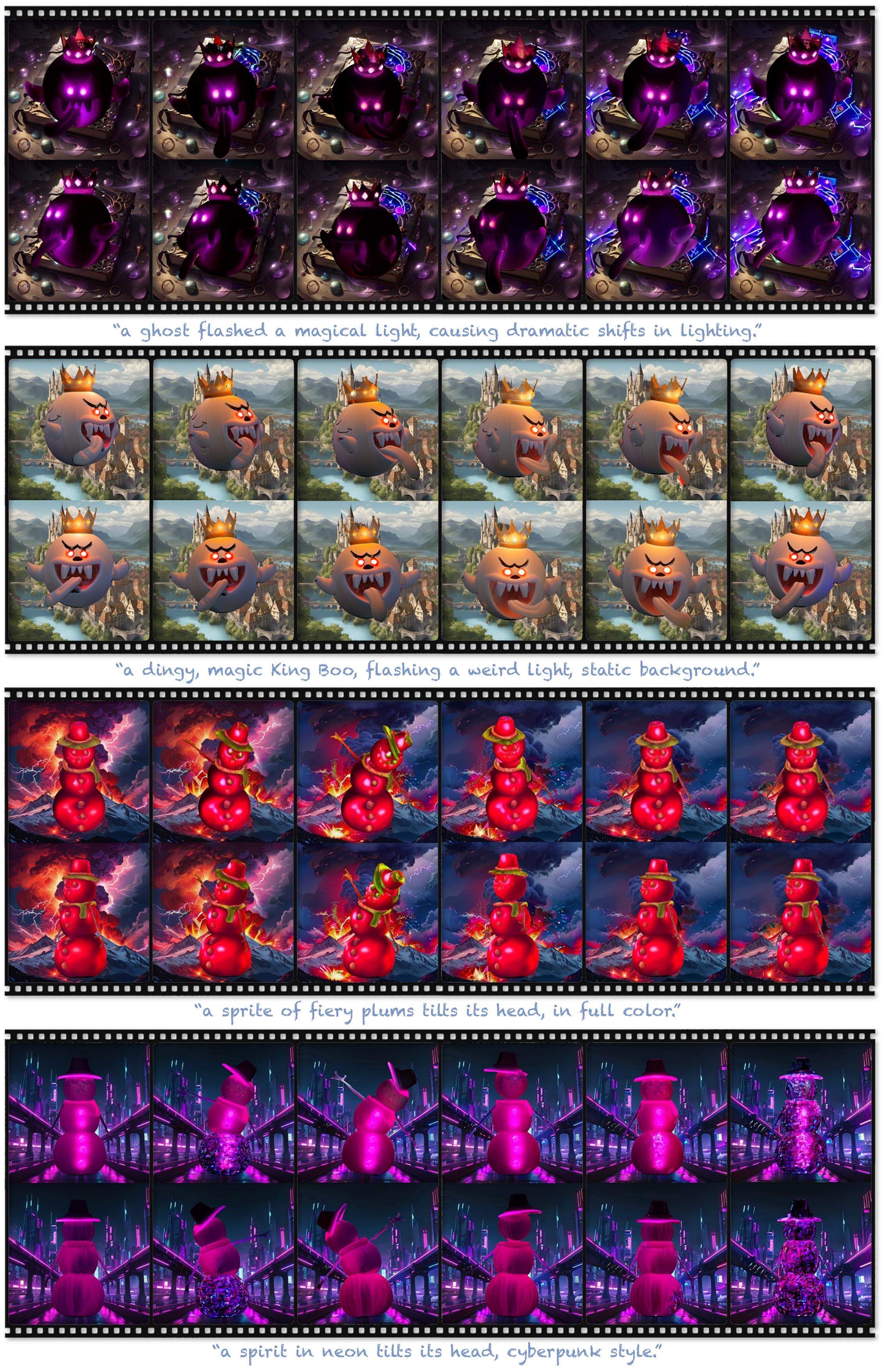}
    \vspace{-1em}
    \caption{\textbf{More qualitative results on non-human character animations.} We present the results of Tex4D with prompts emphasizing the dynamics, demonstrating the ability of Tex4D to capture the dynamics from video diffusion models.}
    \label{fig:supp_results_2}
\end{figure}

\newpage

\subsection{Graphics Application and Video Demo}
\label{sec:supp_demo}

As shown in Fig.~\ref{fig:tex4d_app}, Tex4D demonstrates its utility in the graphics pipeline by integrating dynamic texture sequences into Blender for rendering. This integration enables seamless visualization of animated textures directly on 3D models, showcasing Tex4D's capability to handle complex visual dynamics in real-world applications. We highly recommend the reviewers watch our supplementary videos for details.

\begin{figure*}
    \centering
    \includegraphics[width=\linewidth]{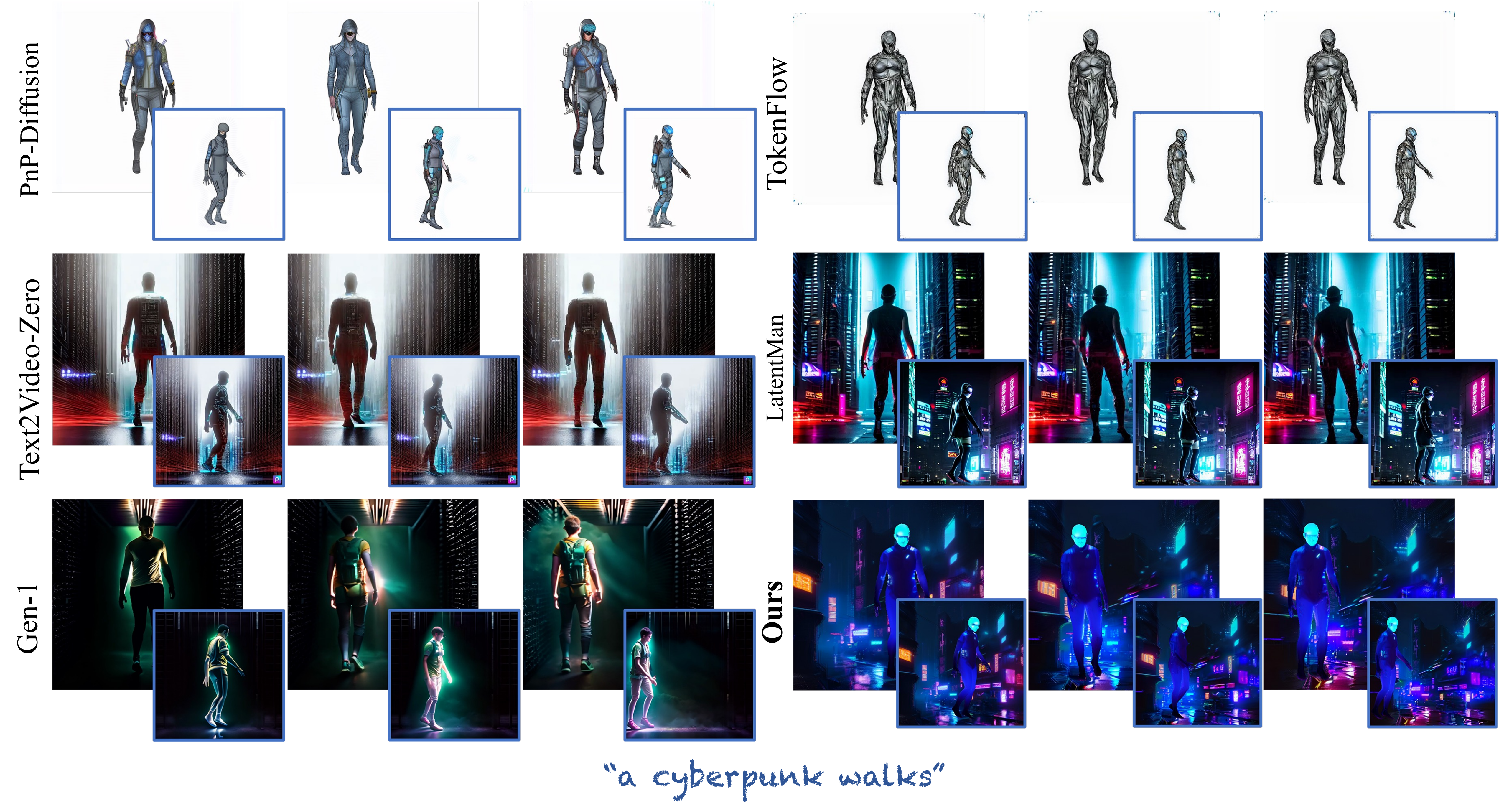}
    \vspace{-1em}
    \caption{\textbf{Qualitative comparisons of multi-view video generation.} We compare our method against PnP-diffusion~\citep{Tumanyan2023pnp}, TokenFlow~\citep{geyer2023tokenflow}, Text2Video-Zero~\citep{text2video-zero}, Generative Rendering~\citep{eldesokey2024latentman}, and Gen-1~\citep{esser2023structure}. We generate videos in the front view and the side view (\textcolor{blue}{blue} box) on Human Diffusion Models. Our method generates vivid videos that align with the textual prompts while preserving spatial consistency.}
    \vspace{-1em}
    \label{fig:comp2}
\end{figure*}

\subsection{Multi-view Results}
\label{sec:supp_multiview}


In this section, we discuss the multi-view results of our methods and other methods. As shown in Fig.~\ref{fig:comp2}, our method generates vivid videos that align with the textual prompts while preserving spatial consistency. PnP-Diffusion~\citep{Tumanyan2023pnp}, TokenFlow~\citep{geyer2023tokenflow} and Text2Video-Zero~\citep{text2video-zero} generate videos that are not aligned with the text prompt, due to the implicit correspondence used in the multi-frame attention. LatentMan~\citep{eldesokey2024latentman} and Gen-1~\citep{esser2023structure} generate vivid videos, but the multi-view consistency of the characters is not well-preserved.

In Fig.~\ref{fig:supp_results}, we present additional characters generated by Tex4D, showcasing the method's effectiveness and its ability to generalize across a diverse array of styles and prompts.
We also evaluate Tex4D on non-human character animations in Fig.~\ref{fig:supp_results_2}, demonstrating its robust generalization capabilities across various types of mesh sequences.
In each case, we provide two different views to show that our method can ensure multi-view consistency.

To emphasize the temporal changes in the generated textures, we also design some prompts, for example, `flashed a magical light', `dramatic shifts in lighting', `cyberpunk style' in our experiments.
As shown in Fig.~\ref{fig:supp_results_2}, the results of `ghost', `King Boo' and `Snowman' validate the effectiveness of our method in generating different level of temporal changes by a variety of textual prompts, while maintaining the consistency both spatially and temporally.
Additionally, we provide a supplementary video that includes baseline comparisons and multi-view results for all examples.

\begin{figure}[t]
    \centering
    \includegraphics[width=\linewidth]{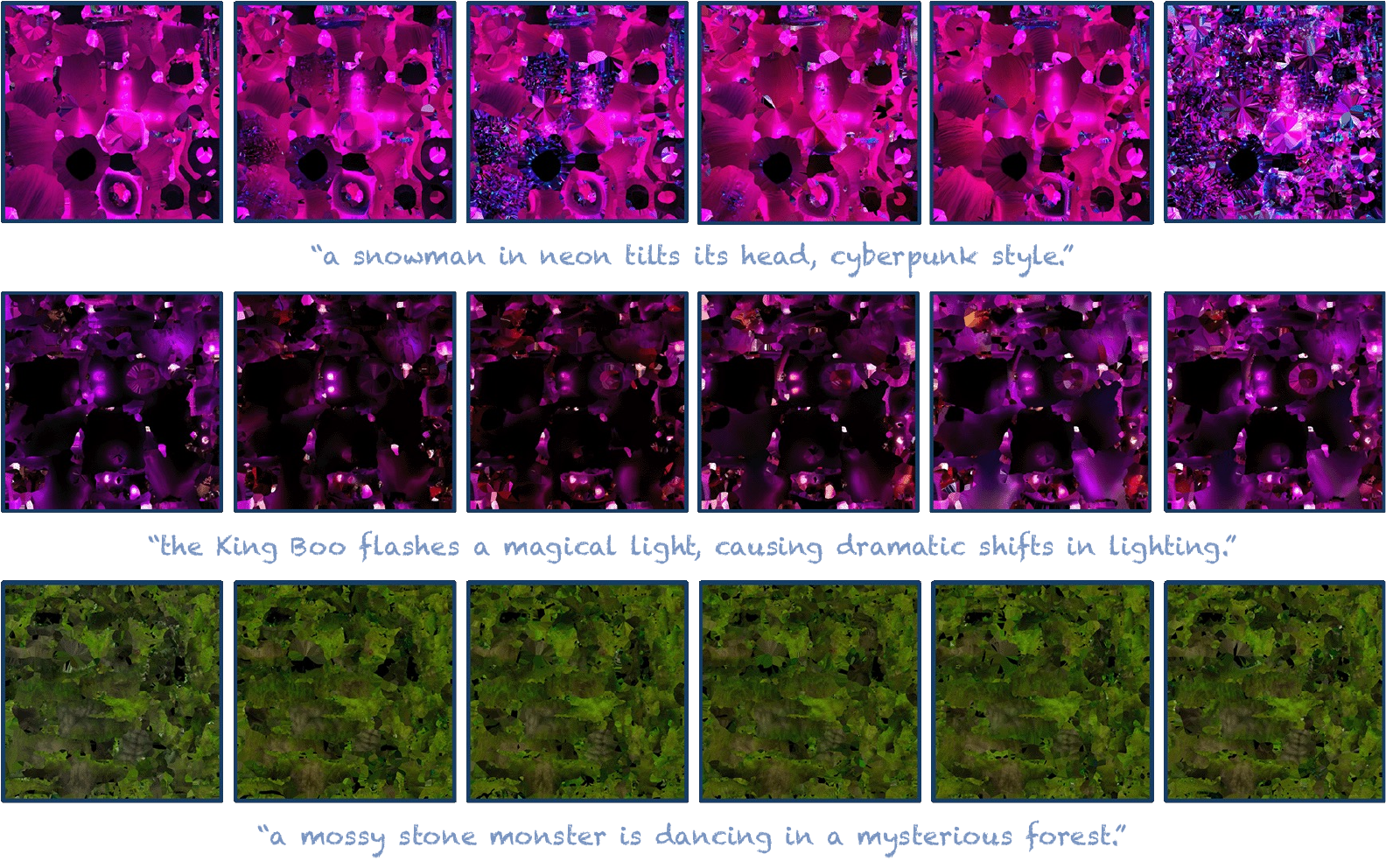}
    \vspace{-1em}
    \caption{\textbf{Visualization of generated textures for mesh sequences.} Our method effectively incorporates temporal changes, such as lighting variations and appearance transformations, directly into the textures, eliminating the need for post-production by technical artists.}
    \label{fig:textures}
    \vspace{-1em}
\end{figure}

\subsection{Texture Results}
\label{sec:supp_texture}

In this section, we present the texture sequences, which are the intermediate results of our pipeline. Our method utilizes XATLAS to unwrap the UV maps from meshes without human labor. XATLAS is a widely used library for mesh parameterization and UV unwrapping, commonly integrated into popular tools and engines, facilitating efficient texture mapping in 3D graphics applications. As shown in~Fig.~\ref{fig:textures}, our method seamlessly bakes temporal changes, including lighting variations, wrinkles, and appearance transformations, directly into the textures, removing the need for manual post-production by technical artists.

\begin{figure}[t]
    \centering
    \includegraphics[width=\linewidth]{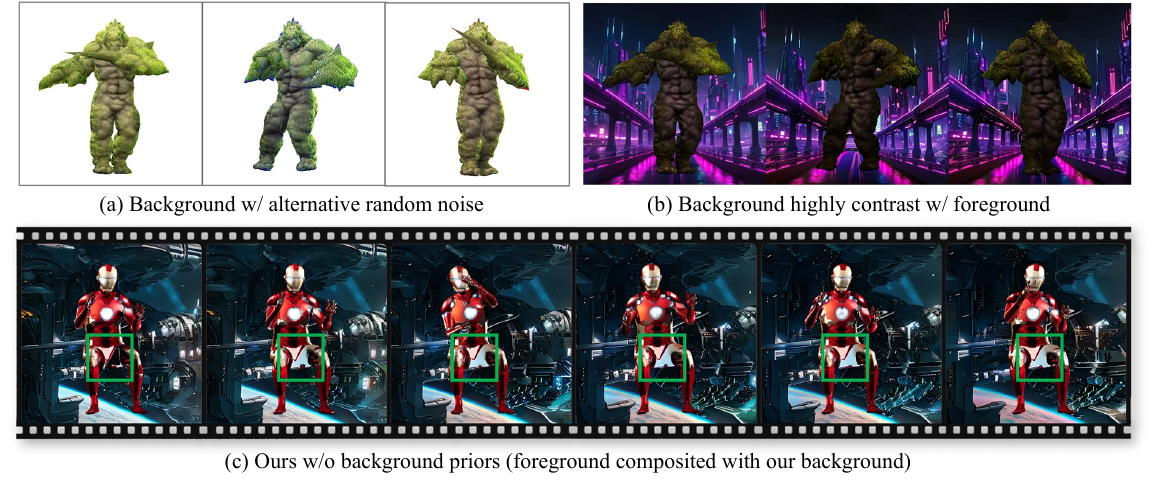}
    \vspace{-1.5em}
    \caption{\textbf{More ablation study on the background priors.} We present three ablations, including the approach used in the texture synthesis method SyncMVD~\citep{liu2023text}, a background that contrasts sharply with the foreground, and without background priors.}
    \label{fig:ablation_bg}
    \vspace{-2em}
\end{figure}

\section{More Ablation Results}

\paragraph{Ablation on Background} To show the effects of various background latent initialization strategies, we provide additional examples, including the approach used in the texture synthesis method~\citep{liu2023text} and a background that contrasts sharply with the foreground object, as shown in~Fig.~\ref{fig:ablation_bg}. In detail, SyncMVD~\citep{liu2023syncdreamer} encodes the backgrounds with alternative random solid color images. For the high-contrast background experiment, we use the latents obtained from the DDIM inversion of highly contrasted foreground and background to initialize the latents.



\section{More Method Comparisons}

\subsection{Comparison with Depth-Conditioned Video Diffusion Models}

\begin{figure}[t]
    \centering
    \includegraphics[width=\linewidth]{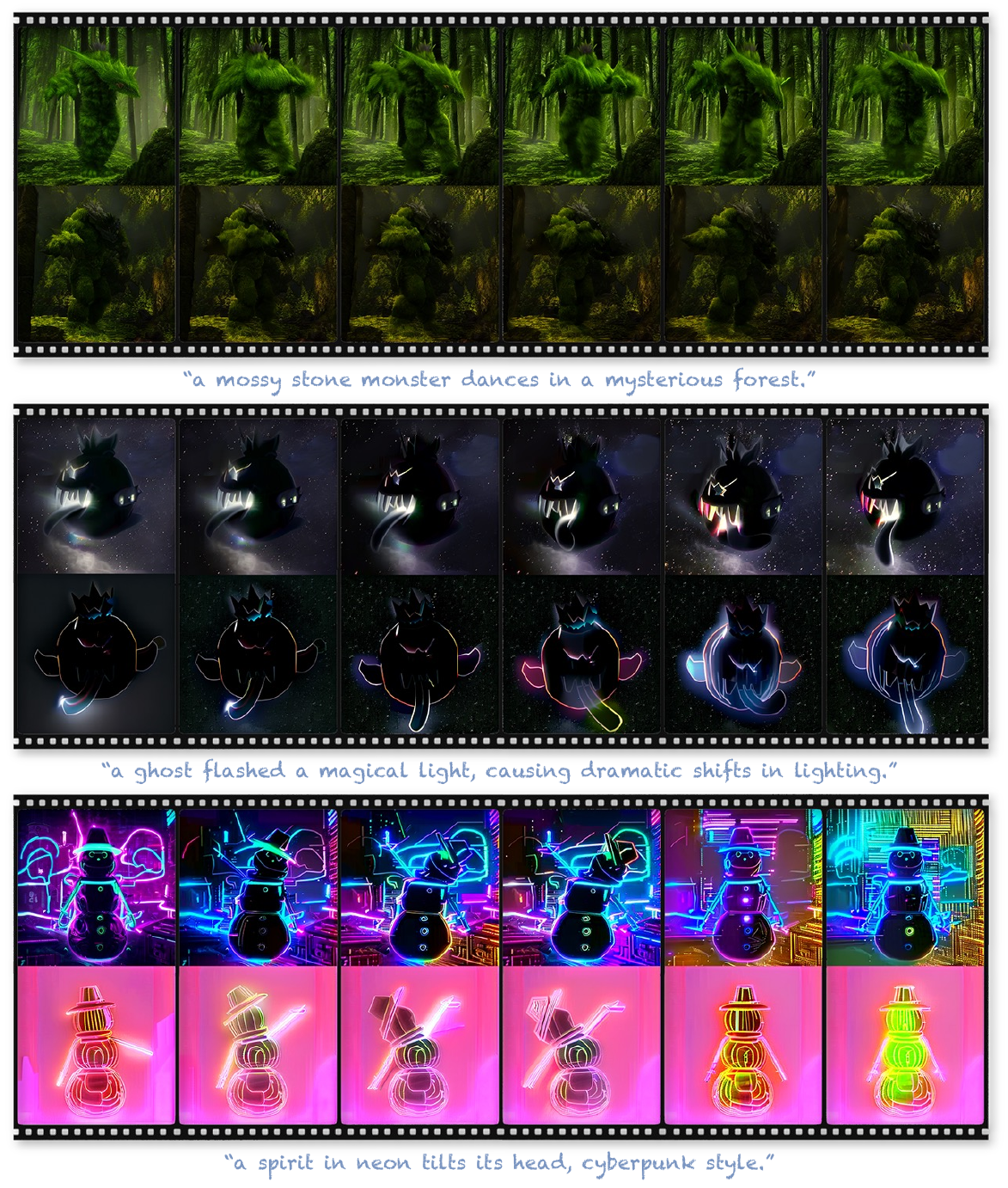}
    \vspace{-1.8em}
    \caption{\textbf{Multi-view results from conditioned video diffusion models.} The conditioned video diffusion models struggle to maintain consistent multi-view representations of a single object due to the depth condition being view-dependent and lacking global geometry information.}
    \label{fig:vdm_results}
    \vspace{-2em}
\end{figure}

While depth-conditioned video diffusion models effectively generate visually compelling results from a single viewpoint, they often struggle to maintain consistent multi-view representations of a single object, such as a character, across different perspectives. To highlight this limitation, we present multi-view results from the depth-conditioned video diffusion model in Fig.~\ref{fig:vdm_results}. The primary cause of this issue is that depth conditions are inherently view-dependent and lack global geometry information, in contrast to UV maps, which provide global information about the 3D space, enabling a unique mapping for each 3D point across all views.

\begin{figure}[t]
    \centering
    \includegraphics[width=\linewidth]{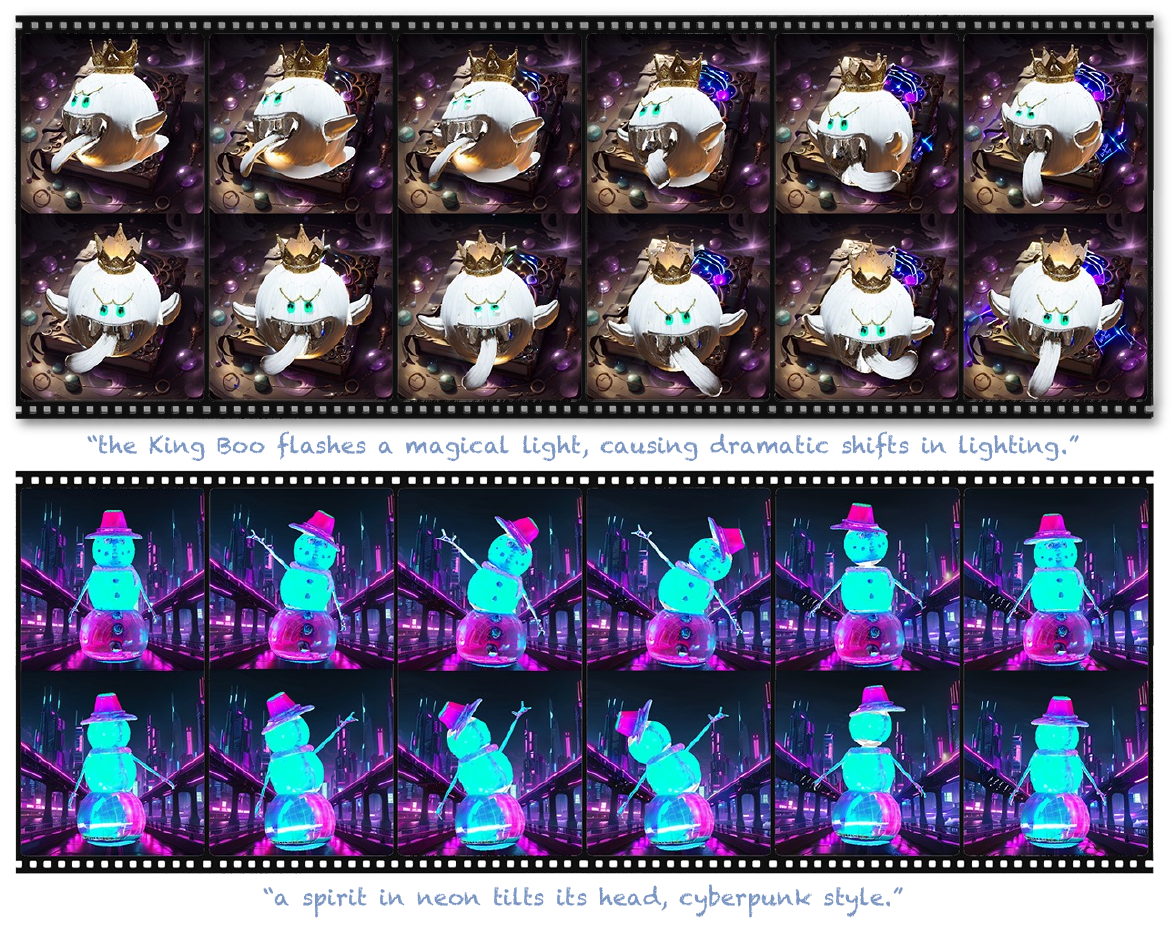}
    \vspace{-1.5em}
    \caption{\textbf{Results of textured mesh animation.} We present the visual results of Text2Tex~\citep{chen2023text2tex} with our backgrounds. Text2Tex fails to capture temporal variations between frames and results in empty texels in invisible regions.}
    \label{fig:text2tex}
    \vspace{-2em}
\end{figure}

\subsection{Comparison with Textured Mesh Animations}

In this section, we highlight the differences between our method and traditional approaches, demonstrating the effectiveness of 4D texturing in capturing temporal variations (e.g., lighting and appearance transformation) within mesh sequences to produce vivid visual results.
Traditional methods typically involve texturing a base mesh (often called a clay mesh) and animating it using skinning techniques. This approach lacks the capacity to represent dynamic visual transformations and the animated sequence is then refined by technical artists who control scene lighting or simulate cloth dynamics to achieve the final visual presentation. This process is labor-intensive and demands specialized expertise in cinematic production and technical engines.

Instead, Tex4D introduces the first solution to dynamic texture generation by leveraging the expressiveness of video diffusion models. This is a fundamental task in filmmaking and game design to model character appearance transformations.

Compared with the static texture generation method, our method presents an alternative by directly integrating complex temporal changes into mesh sequences. As shown in~Fig.~\ref{fig:qualitative},~\ref{fig:supp_results} and~\ref{fig:supp_results_2}, our approach effectively captures temporal effects such as dynamic lighting, and evolving appearances using textual prompts, significantly simplifying the workflow while maintaining high-quality results.

We demonstrate the limitations of traditional textured mesh animation in handling complex temporal changes in~Fig.~\ref{fig:text2tex}. Specifically, we employ the Text2Tex~\citep{chen2023text2tex} to generate the texture for the input mesh in T-pose and render it from multiple viewpoints. To ensure a fair comparison, we composite the rendered results with the background generated by our method.
Notably, the `ghost' and `snowman' examples exhibit visible seams during animation due to self-occlusions that commonly appear in dynamic poses but cannot be accurately predicted during T-pose texture generation. Although the texture is still globally consistent, Text2Tex not just fails to model dynamic effects like appearance transformation but also results in empty texels and disrupts the visual continuity of the animation. For the rendered video results, please kindly refer to our supplementary videos.

\subsection{Comparison with Text-to-4D Methods}

Although the setting of the Text-to-4D task is different from Tex4D as the mesh sequence is not given, we also provide a comprehensive comparison for Text-to-4D methods like SV4D~\cite{xie2024sv4d} and L4GM~\cite{ren2024l4gm} as in Fig.~\ref{fig:comp_t24d}. In our experiments, we found that these Text-to-4D methods usually fail to generate plausible results and have these limitations:
\begin{itemize}[leftmargin=*]
\setlength\itemsep{-.2em}
    \item Multiview video-based methods (e.g., SV4D) struggle with consistency under significant motion. In our early attempts, we also found that the diversity of multiview attention-based inference heavily depends on dataset quality. SV4D handles only simple character animations.
    \item Animatable Gaussian-based methods (DreamGaussian4D, L4GM) built on LGM suffer from blurry and static textures, as decoupling geometry and appearance simultaneously is more challenging.
\end{itemize}

\begin{figure*}
    \centering
    \includegraphics[width=\linewidth]{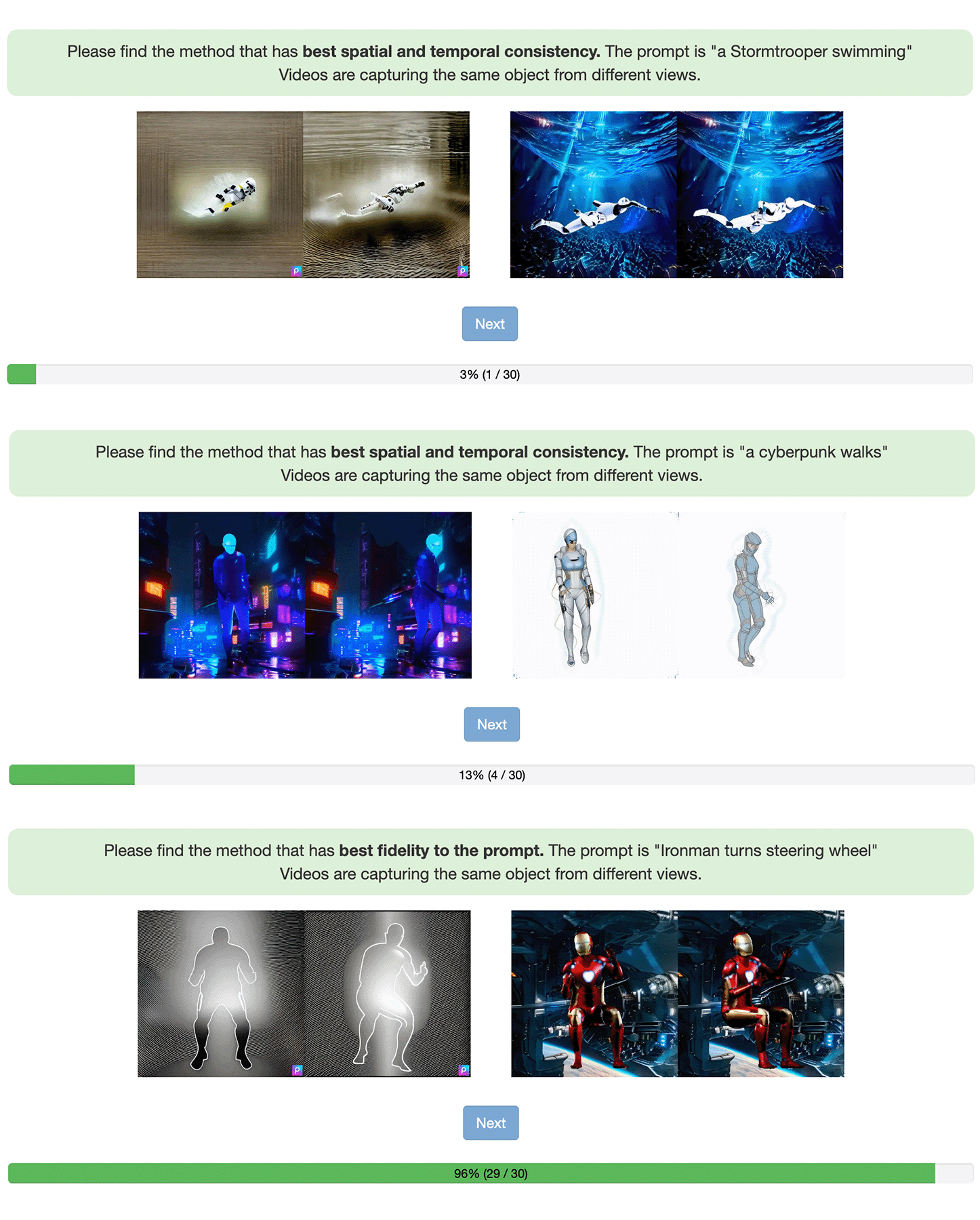}
    \caption{\textbf{User Study.} We provide more visual examples and include quantitative results from our user study. We evaluate the videos from three metrics: Appearance Quality, Spatial and Temporal Consistency, and Fidelity to the Prompt.}
\end{figure*}

\section{User Study}

Our study included 67 participants who provided 1,104 valid responses across six scenes from previous works, each rendered from two different viewpoints. We show each participant 30 pairs of videos synthesized by different methods, capturing the same object from different views.
For each pair, each participant is asked three questions in sequence:
\begin{itemize}[leftmargin=*]
\setlength\itemsep{-.2em}
    \item Which method has better appearance quality?
    \item Which method has better spatial and temporal consistency?
    \item Which method has better fidelity to the prompts?
\end{itemize}

Our study involved 67 participants who provided a total of 1104 valid responses based on six different scenes drawn from six previous works, with each scene producing videos from two different views.

In addition, we further invite 24 participants who provided 455 valid trials to statistically evaluate user preferences (rating from 1 to 5) across Appearance quality, Spatio-temporal consistency, and Prompt Consistency. Users consistently favored Tex4D over all baselines.

\begin{figure}[tp]
    \centering
    \includegraphics[width=\linewidth]{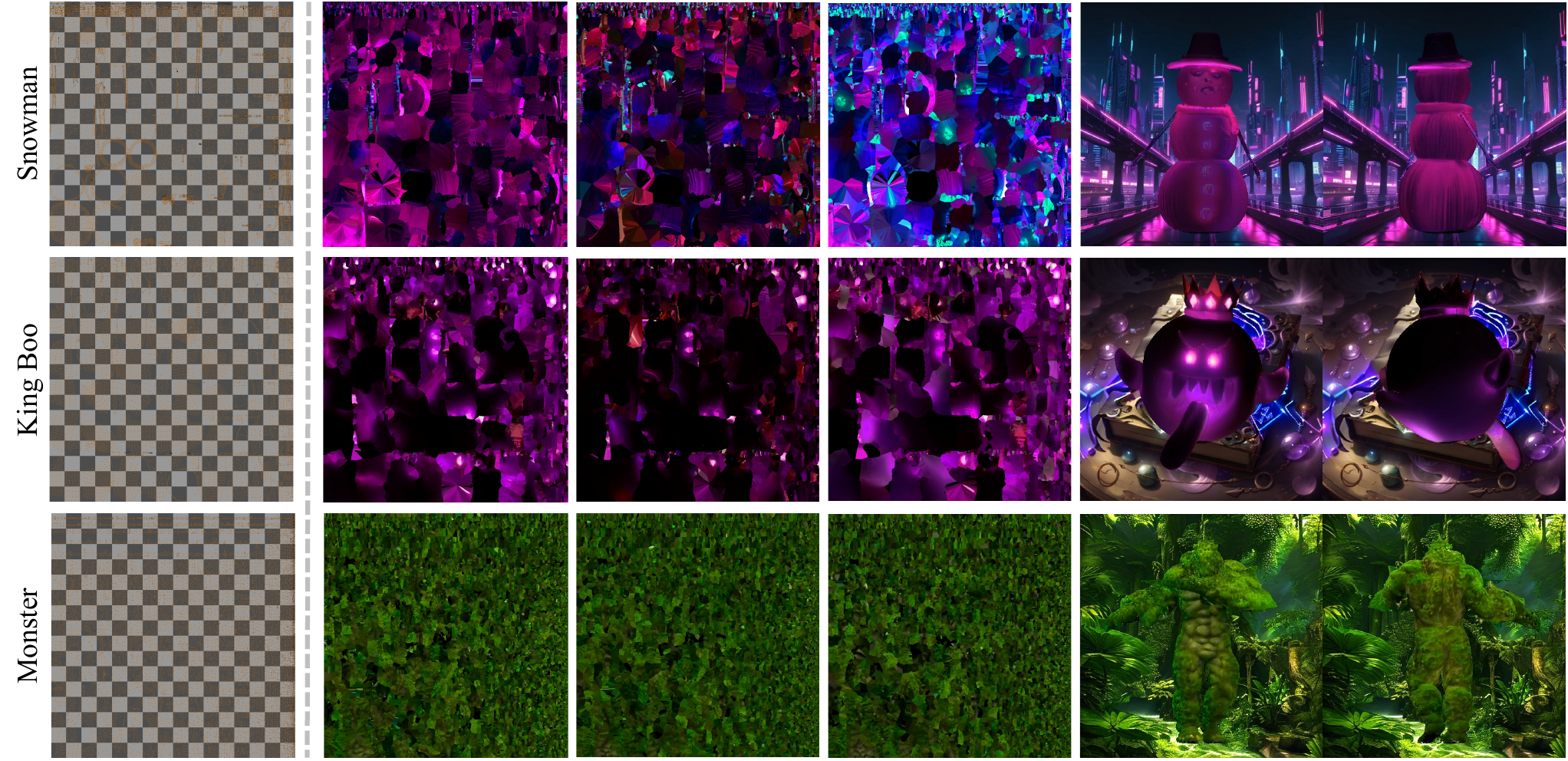}
    \caption{Complex UV with seams created by Blender Smart UV (angle limit=0.5). Tex4D is robust with different UV mappings. First column: UV without texels. Last column: rendered video frames. Other columns: texture generated by Tex4D. Zoom in for texture details.}
    \label{fig:robust}
\end{figure}

\section{Broader Discussion}

As Tex4D is the first work to generate dynamic textures, we provide a broader analysis of our method in this section and hope to provide more insights for future works.

\subsection{Robustness of Different UV Mappings}

To validate the robustness of our method in different UV mappings, we test Tex4D using Blender’s smart UV strategy. Specifically, we set the angle limit for adjacent face normals to 30 degrees for splitting with more seams. The rendered results are also visually robust as shown in Fig~\ref{fig:robust}.

\begin{figure}[htp]
    \centering
    \includegraphics[width=\linewidth]{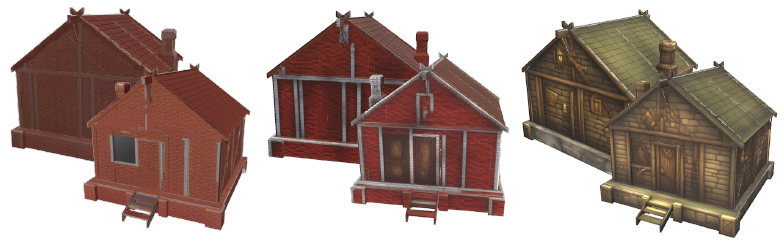}
    \caption{\textbf{Highly structured texture generation.} A low-poly house is created with different prompts to demonstrate the capability of Tex4D to generate diverse, highly structured textures. Zoom in for texture details.}
    \label{fig:supp_structured}
\end{figure}

\subsection{Highly Structured Texture Generation}

To further demonstrate the robustness of our method, we test Tex4D with the task of highly structured texture generation that is commonly used in relief mapping and displacement mapping. We create a low-poly house as the base mesh and test our method with different prompts as shown in~Fig. \ref{fig:supp_structured}. Tex4D could generate multi-view consistent and highly structured texture, demonstrating our method is robust across different tasks.

\subsection{Runtime Breakdown}

\begin{wraptable}{r}{0.52\textwidth}
    \vspace{-5mm}
    \centering
\resizebox{0.5\textwidth}{!}{
    \begin{tabular}{l|cccc}
    \toprule
    \textbf{Keyframes} & \textbf{3} & \textbf{10} & \textbf{17} & \textbf{25} \\
    \midrule
    Time (min) w/ res. (96×96) & 4.1 & 13.6 & 26.6 & 34.4 \\
    Time (min) w/ res. (64×64) & 2.3 & 6.8  & 10.7 & 14.4 \\
    \bottomrule
    \end{tabular}
}
    \caption{\textbf{Quantitative results of runtime breakdown.}}
    \label{tab:runtime}
    \vspace{-3mm}
\end{wraptable}

In~Tab. \ref{tab:runtime}, we provide a runtime breakdown including different keyframe numbers and latent resolutions. We have tested the latent resolutions with $96\times96$ and $64\times64$, and the keyframe numbers $3$, $10$, $17$ and $25$ respectively in the video generation task with a total frame number of 72. In our default setting, we use the keyframe interval of 3 and latent size $96\times96$ for better generation quality.

\subsection{Long Texture Sequence Generation}

\begin{figure*}[tp]
    \centering
    \includegraphics[width=\linewidth]{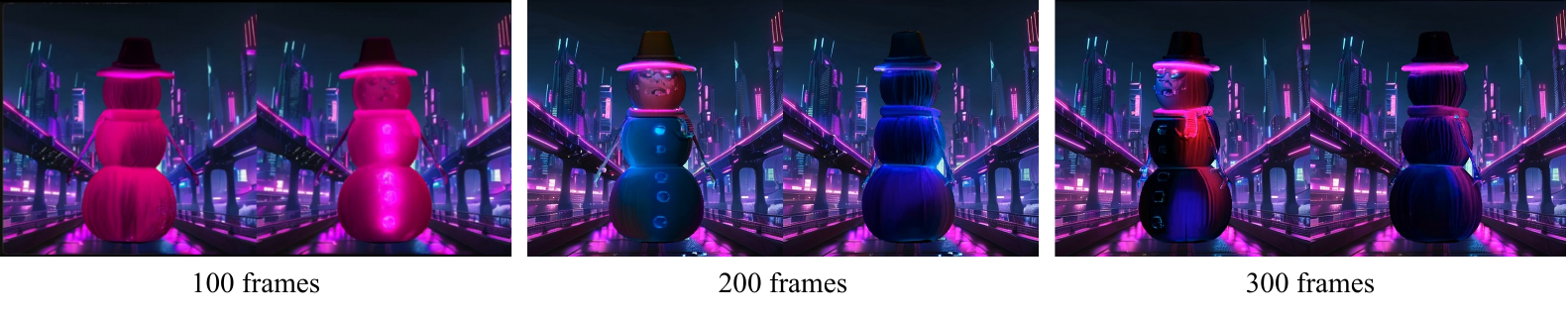}
    \caption{\textbf{Long Texture Sequence Generation.} Our method could maintain plausible textures with the increase in animation sequences. However, the quality may degrade due to the training sets of video diffusion models usually do not contain such long sequences.}
    \label{fig:supp_longseq}
\end{figure*}

In this section, we extend the original mesh sequence to longer lengths to study the robustness of Tex4D at long texture sequence generation. Specifically, we use the ``snowman'' case, which originally has 100 frames. We extend the mesh sequence to 200 and 300 frames by repeating the animation. Tex4D is robust in generating long sequence textures while preserving the dynamics. We provide the result of intermediate frames in~Fig. \ref{fig:supp_longseq}. However, the texture details may degrade due to the video diffusion models, which tend to distort high-frequency components in long videos, as indicated by~\citep{lu2024freelong}. Moreover, the training sets of video diffusion models usually do not contain such long sequences, which hinders the model to learn long sequence appearance as indicated by~\citep{chen2024videocrafter2}.

\subsection{Analysis of Appearance Details}

During our early experiments and literature review, we found video diffusion models tend to generate relatively smooth results compared with image diffusion models, as indicated by DiffusionRenderer~\cite{liang2025diffusionrenderer} and I2VGen-XL~\citep{2023i2vgenxl}. As a result, in some cases, the clown's face lacks fine details within small pixel patches as shown in Fig.~\ref{fig:ablation_blend}.

We analyze the underlying causes from three perspectives.
Intuitively, maintaining high-frequency details temporally is harder than a smooth one, and video diffusion models tend to comprise the spatial details for better temporal consistency.
Technically, this problem should be attributed to the data limitations, including dataset scales and low-quality patches from websites (e.g., WebVid10M~\citep{Bain21web}) used for video diffusion models, as analyzed by VideoCrafter2~\citep{chen2024videocrafter2}.
Specifically for Tex4D, the latents aggregation module (Sec.~\ref{sec:texaggr}) may potentially wipe out details because of the average operation on latents, although with the weighted cosine maps. A possible step is to train a tiny adapter like TexFusion~\cite{cao2023texfusion} to optimize the latents aggregation instead of RGB mapping.

Some concurrent works attempt to address this problem, either from the data distribution view (FreqPrior~\cite{yuan2025freqprior}) or the U-Net architecture view (CogVideoX~\citep{yang2024cogvideox}). As our methods can be easily integrated into any video diffusion model, we anticipate improving the quality of Tex4D with the advancement of video diffusion models.

\subsection{Societal Impacts}
\label{sec:social_impact}

\textbf{Potential Social Impacts} Although our method offers broad potential for controllable generation tasks, like other frameworks for image and video synthesis, it could be misused for harmful purposes (such as generating deceptive content or fake media). As such, responsible and cautious use is essential when applying it in real-world scenarios.

\textbf{Safeguards} During inference, we enable the NSFW filter provided by the underlying models to help prevent the generation of explicit or inappropriate content, thereby safeguarding users from unwanted exposure. The training datasets used in our base model, CTRL-Adapter, have already filter out the image/video samples with harmful contents. Specifically, Panda70M filters harmful content using an automated pipeline and replaces names with ``person'' via NLTK. Similarly, LAION-POP applies a custom NSFW classifier to exclude unsafe samples.

\section{Limitations}
\label{sec:supp_limit}

\subsection{Panoramic Background Modeling}

One limitation of our method is the lack of seamless integration between the generated textures and the background, resulting in a disjointed appearance where the foreground and background elements may seem artificially stitched together.
However, the dynamic textures remain globally consistent and can be directly applied to the downstream tasks, as shown in Fig.~\ref{fig:tex4d_app}.
To the best of our knowledge, no existing work tackles the foreground and background texture generation together because the task is computationally expensive, and the scene-level dataset is limited.
Addressing the scene-level 4D texturing remains an open challenge that we aim to explore in future work. 

\subsection{Computation Time}
We notice that our method is relatively computationally intensive compared with other texture synthesis methods. The running time of our method primarily depends on the foundation model CTRL-Adapter, which takes approximately 5 minutes to generate a 24-frame video. We anticipate efficiency improvements with advancements in conditioned video diffusion models to further enhance the practicality of Tex4D.

\section{Licenses}

We use standard licenses from the community
and provide the following links to the licenses for the datasets, codes, and models that we used in this
paper. For further information, please refer to the specific link.

PyTorch~\cite{Ansel_PyTorch_2_Faster_2024}: \href{https://github.com/pytorch/pytorch/blob/main/LICENSE}{BSD-style}

HuggingFace Transformers~\cite{wolf-etal-2020-transformers}: \href{https://github.com/huggingface/transformers/blob/main/LICENSE}{Apache License 2.0}

HuggingFace Diffusers~\cite{von-platen-etal-2022-diffusers}: \href{https://github.com/huggingface/diffusers/blob/main/LICENSE}{Apache License 2.0}

CTRL-Adapter~\citep{lin2024ctrladapter}: \href{https://github.com/huggingface/diffusers/blob/main/LICENSE}{Apache License 2.0}

ControlNet~\cite{zhang2023adding}: \href{https://github.com/lllyasviel/ControlNet?tab=Apache-2.0-1-ov-file}{Apache License 2.0}

I2VGen-XL~\cite{zhang2023i2vgen}: \href{https://huggingface.co/ali-vilab/i2vgen-xl}{MiT License}



\end{document}